\documentclass[twoside]{article}
\usepackage[preprint]{aistats2026}
\usepackage{graphicx}
\usepackage{algorithm}
\usepackage{algorithmic}
\usepackage{amsmath}
\usepackage{amssymb}
\usepackage{tabularx}
\usepackage{booktabs}
\usepackage{multirow}
\usepackage{amsthm}
\usepackage{subfigure}
\usepackage{caption}
\usepackage{array}
\usepackage{url}

\usepackage[round]{natbib}

\newtheorem{theorem}{Theorem}
\newlength{\mycolsep}
\begin{document}

%

%
\runningtitle{Dualformer: Time-Frequency Dual Domain Learning for Long-term Time Series Forecasting}

\runningauthor{Jingjing Bai and Yoshinobu Kawahara}

\twocolumn[

\aistatstitle{Dualformer: Time-Frequency Dual Domain Learning for\\ Long-term Time Series Forecasting}

\aistatsauthor{ Jingjing Bai \And Yoshinobu Kawahara*}

\aistatsaddress{ The University of Osaka \& RIKEN AIP\\jingjing.bai@ist.osaka-u.ac.jp \And  The University of Osaka \& RIKEN AIP \\kawahara@ist.osaka-u.ac.jp } ]

\begin{abstract}
Transformer-based models, despite their promise for long-term time series forecasting (LTSF), suffer from an inherent low-pass filtering effect that limits their effectiveness. This issue arises due to undifferentiated propagation of frequency components across layers, causing a progressive attenuation of high-frequency information crucial for capturing fine-grained temporal variations. To address this limitation, we propose Dualformer, a principled dual-domain framework that rethinks frequency modeling from a layer-wise perspective. Dualformer introduces three key components: (1) a dual-branch architecture that concurrently models complementary temporal patterns in both time and frequency domains; (2) a hierarchical frequency sampling module that allocates distinct frequency bands to different layers, preserving high-frequency details in lower layers while modeling low-frequency trends in deeper layers; and (3) a periodicity-aware weighting mechanism that dynamically balances contributions from the dual branches based on the harmonic energy ratio of inputs, supported theoretically by a derived lower bound. This design enables structured frequency modeling and adaptive integration of time-frequency features, effectively preserving high-frequency information and enhancing generalization. Extensive experiments conducted on eight widely used benchmarks demonstrate Dualformer’s robustness and superior performance, particularly on heterogeneous or weakly periodic data. Our code is publicly available at \url{https://github.com/Akira-221/Dualformer}.
\end{abstract}

\section{Introduction}

Long-term time series forecasting (LTSF) has attracted increasing attention due to its wide applicability in domains such as finance~\citep{ariyo2014stock}, healthcare~\citep{alaa2019attentive},energy~\citep{bilal2022comparative}, and climate modeling~\citep{duchon2012time}. Recent deep learning models, particularly Transformer-based architectures~\citep{liu2023itransformer,wu2021autoformer,zhou2021informer, liu2022pyraformer}, have shown strong ability to model long-range dependencies. To further enhance global pattern learning, several state-of-the-art models, including FEDformer~\citep{zhou2022fedformer}, TimesNet~\citep{wu2022timesnet}, and PDF~\citep{dai2024periodicity}, introduce frequency-enhanced components and have achieved impressive performance on various benchmarks.

Despite these advances, the effectiveness of these models in capturing informative temporal variations remains limited. A fundamental reason lies in the inherent frequency bias of the self-attention mechanism. Theoretical analyses~\citep{dong2021attention,zhang-etal-2021-orthogonality} have shown that deep self-attention networks suffer from rank collapse and diminished representation diversity due to exponentially decaying attention distributions. Recent studies~\citep{wang2022anti,zhang2024not} further revealed that self-attention inherently acts as a low-pass filter, progressively suppressing high-frequency components. These high-frequency components, often representing rapid changes or short-term fluctuations, are critical for accurate forecasting in real-world scenarios like weather, finance, or energy. Mitigating this frequency bias is therefore key to leveraging both local and global temporal patterns.

We argue that this low-pass bias of Transformers arises from their uniform propagation of all frequency components across layers, allowing higher-energy, low-frequency patterns (e.g., trends and seasonality) to naturally dominate the attention scores. As the network deepens, this imbalance intensifies, leading to the gradual fading of lower-energy, high-frequency signals. To address this, we propose to explicitly restructure the frequency propagation path within the model. Inspired by the hierarchical nature of neural representation learning~\citep{zeiler2013visualizingunderstandingconvolutionalnetworks, tenney2019bertrediscoversclassicalnlp}, we introduce a layer-wise frequency decomposition strategy that allocates high-frequency components to lower layers and low-frequency components to deeper layers. This design reflects the fact that shallow layers are more suited for capturing localized, fast-varying patterns, while deeper layers progressively aggregate broader context to model long-term trends. By explicitly controlling frequency flow across layers, our model prevents low-frequency dominance and preserves fine-grained signals that would otherwise vanish in deeper stages.


Based on these insights, we propose Dualformer, a dual-domain Transformer framework that models time and frequency features in parallel and mitigates frequency degradation through structural design. It consists of three key components: (1) a dual-branch architecture that jointly learns temporal and spectral features, capturing complementary information for more expressive and adaptive representations, thereby overcoming the limitations of single-domain approaches; (2) a hierarchical frequency sampling module that structurally addresses high-frequency degradation by assigning distinct frequency bands to different layers, allocating high-frequency components to shallow layers and low-frequency trends to deeper ones, to enable effective multi-scale and multi-resolution feature extraction; and (3) a periodicity-aware weighting mechanism that adaptively fuses the two branches based on input's periodic level measured by its harmonic energy ratio, allowing the model to generalize well to time series with diverse spectral characteristics. Together, these designs allow Dualformer to dynamically integrate information across domains and frequencies, enhancing both accuracy and generalization.

\begin{figure*}[t]
    \centering
    \includegraphics[width=0.9\textwidth]{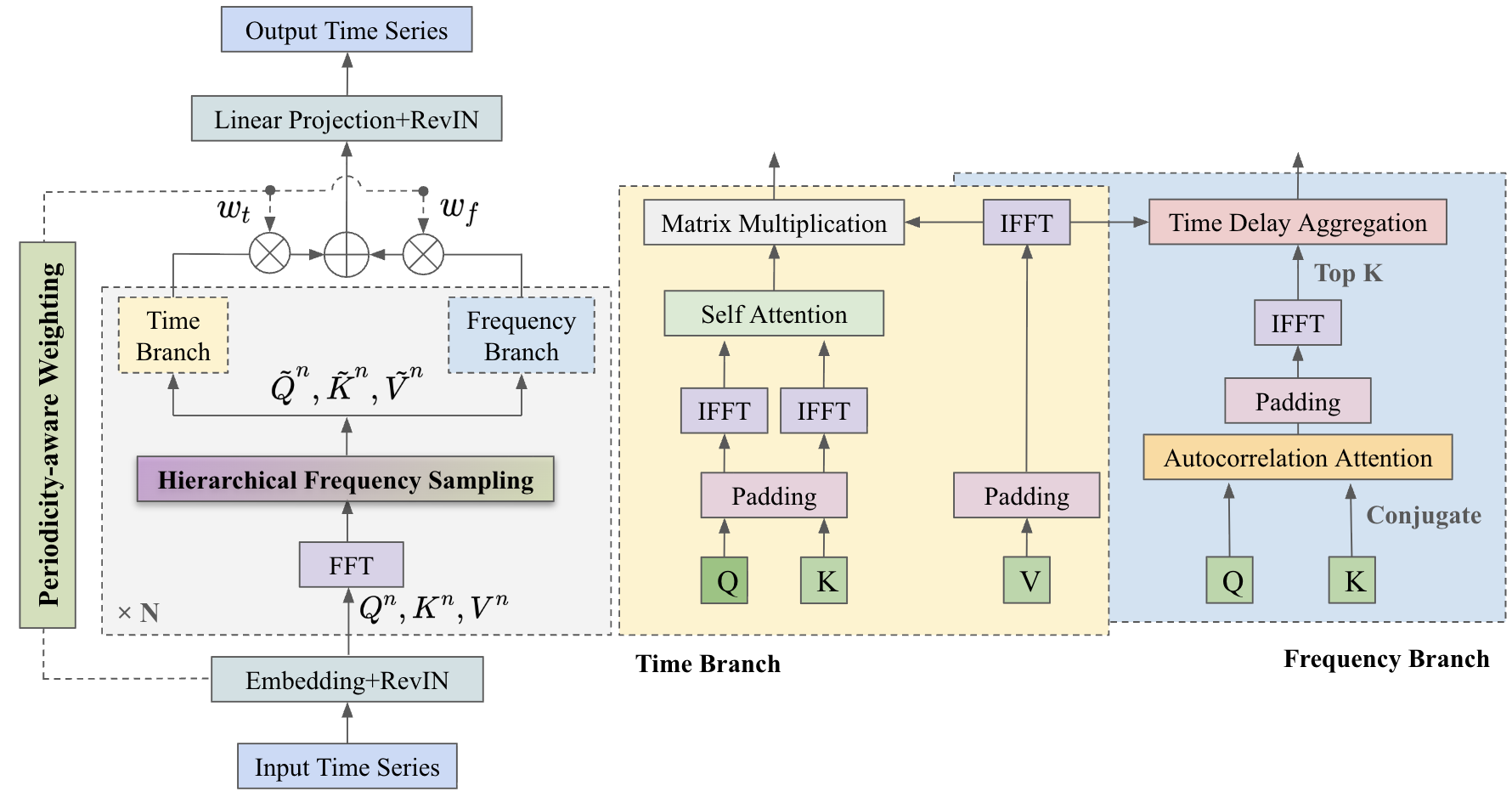}
    \caption{\centering The overall architecture of our proposed Dualformer model.}
    \label{fig:01}
\end{figure*}
We validated Dualformer through extensive experiments on eight benchmark datasets. The results show that Dualformer consistently outperforms existing models, especially on heterogeneous or weakly periodic data.

\section{Related Work}
Transformer-based models have achieved notable success in long-term time series forecasting by leveraging self-attention to model long-range dependencies. Early advances such as Autoformer~\citep{wu2021autoformer} introduced seasonal-trend decomposition and autocorrelation mechanisms to enhance temporal structure modeling, while PatchTST~\citep{nie2022time} improved efficiency through patch-level tokenization. Interestingly, even simple linear models like DLinear~\citep{zeng2023transformers} have demonstrated strong performance, suggesting that complex attention mechanisms can sometimes overfit. However, most of these methods operate purely in the time domain and often ignore the underlying frequency characteristics, leading to the well-documented low-pass filtering effect where high-frequency signals are progressively attenuated.

To mitigate this issue, several frequency-domain methods have been proposed. FEDformer~\citep{zhou2022fedformer} replaced self-attention with a Fourier-based attention to capture global frequency structure, while FiLM~\citep{zhou2022film} combined Fourier and Legendre projections to improve robustness. Others, like FreTS~\citep{yi2024frequency}, simplify spectral modeling with frequency-wise MLPs. Although these methods effectively extract global structure, they typically apply frequency modeling uniformly across layers, lacking a progressive, resolution-aware processing. 

Hybrid approaches like~\cite{ye2024atfnet, wu2022timesnet, yue2025freeformerfrequencyenhancedtransformer, 10242151,kui2025tfkantimefrequencykanlongterm, LUO2025111412, dai2024periodicity} have emerged to combine time and frequency information. TFKAN~\citep{kui2025tfkantimefrequencykanlongterm} and TFDNet~\citep{LUO2025111412} introduced joint time and frequency modeling architectures, while TimesNet~\citep{wu2022timesnet} and PDF~\citep{dai2024periodicity} reshaped the 1D series into a 2D tensor using FFT to capture both long and short-term variations. Although these models incorporated richer structures, they often rely on static or fixed domain integration schemes. Without adapting to the specific spectral characteristics of each input, their generalization ability can be limited on heterogeneous or non-stationary series.

Motivated by these challenges, our work proposes a structured and input-adaptive approach to time-frequency modeling. In contrast to prior methods with uniform processing or static fusion, our model explicitly aligns frequency components with network depth and dynamically adjusts its domain reliance based on the input's spectral properties.


\section{Proposed Model}
The overall architecture of Dualformer is illustrated in Figure \ref{fig:01}. Given an input time series $X \in \mathbb{R}^{L \times C}$, where $C$ is the number of variables, and $L$ is the input sequence length, it is first processed by an embedding layer and reversible instance normalization (RevIN)~\citep{kim2021reversible}. The resulting representation is then transformed into the frequency domain using Fast Fourier Transform (FFT), yielding a complex-valued frequency spectrum. Subsequently, to address the imbalance in spectral modeling, a hierarchical frequency sampling (HFS) module dynamically selects subsets of frequency components across layers, ensuring that lower layers focus on high-frequency features while higher layers progressively shift attention to low-frequency features.

The sampled frequency features are then processed in parallel by a time branch and a frequency branch. In the time branch, the selected frequency components are first converted back to the time domain using inverse FFT (IFFT), followed by a vanilla self-attention that models temporal dependencies and generates time-domain representations. In contrast, the frequency branch applies an autocorrelation-based attention directly in the frequency domain to measure periodic dependencies and global patterns. The resulting correlations are padded, inverse transformed, and subjected to a top-$k$ selection to retain dominant periodic components via time delay aggregation.

The outputs of the two branches are adaptively fused through a periodicity-aware weighting algorithm, which computes a soft weight based on the input's periodicity level. This allows the model to dynamically balance the contributions of time-domain and frequency-domain features. The fused features are finally passed through a linear projection and re-normalized to produce the output time series.

Detailed formulations and design choices for each module are presented in the following sections.

\subsection{Hierarchical frequency sampling}\label{subsec3.1}
To structurally mitigate the low-frequency bias inherent in Transformer-based models, we propose a hierarchical frequency sampling (HFS) strategy that explicitly decomposes the spectrum across model depth. Instead of propagating all frequencies through each layer, HFS assigns high-frequency components to lower layers and low-frequency components to deeper layers, thereby counteracting the implicit low-pass filtering effect of self-attention.

Formally, for the $n$-th layer input $X^n \in \mathbb{R}^{L \times C}$, we apply FFT along the temporal axis of each channel to obtain its spectral representation:
\begin{equation*}
    \mathcal{F}(X^n)\rightarrow W^n\in\mathbb{C}^{M\times C},
\end{equation*}
where $M = \lceil L / 2 \rceil + 1$ due to the conjugate symmetry property of FFT~\citep{brigham1988fast}. We then apply the following layer-specific sampling function $\text{Sampling}^n_\alpha$ to select frequency ranges for the given layer:
\begin{equation*}
    \tilde{W}^n=\text{Sampling}^n_\alpha(W^n)=W[p^n:q^n,:],
\end{equation*}
where $\tilde{W}^n \in \mathbb{C}^{F \times C}$ represents the sampled frequency components for layer $n$, and $F$ denotes the range of the frequency features. The sampling ratio $\alpha = \frac{F}{M}$ is a tunable hyperparameter, which controls the frequency bandwidth allocated to each layer.

\begin{figure}[t]
    \centering
    \includegraphics[width=0.7\linewidth]{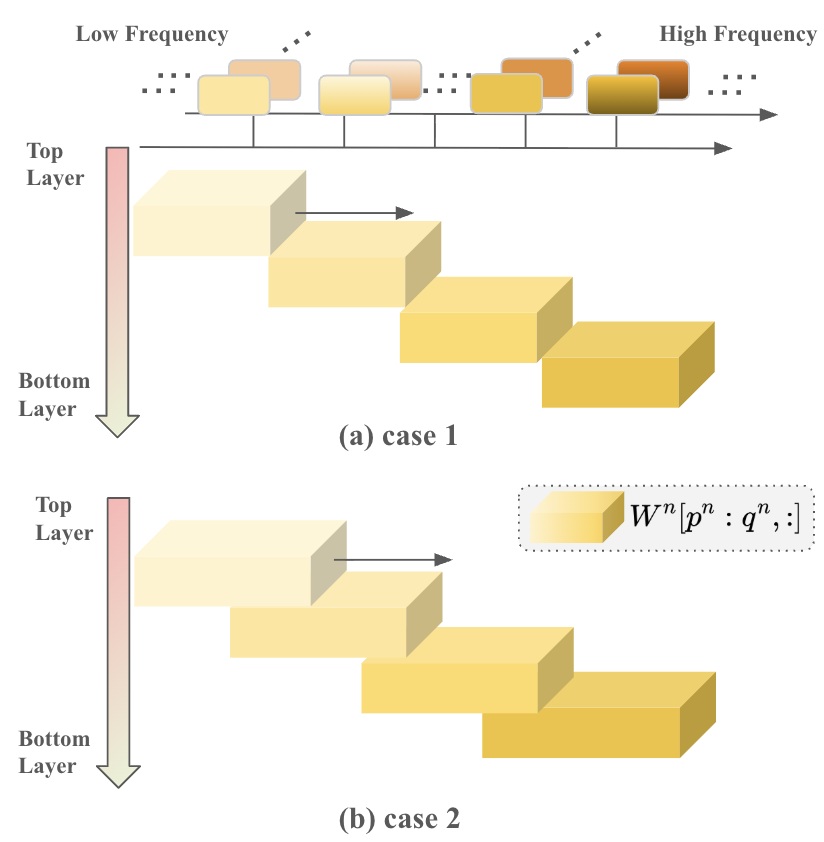}
    \caption{Hierarchical frequency sampling. When $\alpha\le 1/N$, case 1 is adopted; otherwise, case 2 is adopted. }
    \label{fig:02}
\end{figure}

The sampling intervals $p^n$ and $q^n$ are dynamically adjusted based on the layer index $n$, defined as:
\begin{equation*}
    p^n=M(1-\alpha)(1-\frac{n-1}{N-1}),\quad q^n=p^n+\alpha M,
\end{equation*}
where $N$ is the number of encoder layers. This design ensures that as $n$ increases, the sampling window shifts from high to low frequencies. Additionally, the overlap between adjacent layers $n$ and $n+1$ is:
\begin{equation*}
    q^{n+1}-p^n=(\alpha N-1)\frac{M}{N-1}.
\end{equation*}

Therefore, when $\alpha > 1 / N$, the frequency ranges between layers overlap, as shown in case 2 of Figure \ref{fig:02}. While $\alpha < 1 / N$, frequency loss occurs between layers. To prevent such information gaps, we enforce uniform partitioning, as shown in case 1 of Figure \ref{fig:02}. Correspondingly, the updated formulas of $p^n$ and $q^n$ are:
\begin{equation*}
    p^n=M(1-\frac{n}{N}),\quad q^n=p^n+\frac{M}{N}.
\end{equation*}

This hierarchical frequency sampling allows each layer to focus on a distinct frequency range, enabling the model to learn multi-resolution representations while avoiding frequency redundancy or loss. It provides a principled and structurally grounded solution to high-frequency degradation.

\subsection{Time branch}\label{subsec3.2}
As shown in Figure \ref{fig:01}, the time branch is designed to capture local temporal dependencies by combining frequency-guided feature selection with time-domain self-attention. Given the embedded input $X^n$, we apply linear projections to obtain $Q^n, K^n, V^n$, which are first transformed into the frequency domain via FFT. Then, our hierarchical frequency sampling (HFS) module extracts a layer-specific spectral subset $\tilde{Q}^n, \tilde{K}^n, \tilde{V}^n$ to emphasize high-frequency components in shallow layers and low-frequency trends in deeper layers. To enable time-domain modeling, the sampled spectra are padded and reconstructed using inverse FFT:
\[
\tilde{Q}_t^n, \tilde{K}_t^n, \tilde{V}_t^n = \mathcal{F}^{-1}(\text{Padding}(\tilde{Q}^n, \tilde{K}^n, \tilde{V}^n)).
\]

A detailed explanation and pseudocode of the padding operation are provided in the supplementary material. Vanilla self-attention is then performed on these reconstructed features to capture contextual interactions between time steps:
\[
\text{Attention}(\tilde{Q}^n_t,\tilde{K}^n_t,\tilde{V}^n_t)=\text{Softmax}(\frac{\tilde{Q}^n_t\cdot(\tilde{K}^n_t)^\top}{\sqrt{d_k}})\tilde{V}^n_t.
\]

We adopt the standard multi-head attention, where the \emph{query}, \emph{key}, and \emph{value} of the $i$-th head are:
\[
Q_i=W^Q_i Q, K_i=W^K_i K, V_i=W^V_i V, Q_i,K_i, V_i\in \mathbb{R}^{L\times \frac{D}{h}},
\]
where $D$ is the hidden variable dimension and $h$ is the number of heads. The outputs of multiple heads are concatenated and projected as:
\[
\text{Multi-head}(Q,K,V)=\text{Concat}(\text{head}_1,\dots,\text{head}_h)W^o_t,
\]
where $
\text{head}_i=\text{Attention}(Q_i,K_i,V_i)
$.

In essence, the time branch is not a generic temporal feature extractor but a scale-specific dynamics modeler. Our model design breaks down the complex task of modeling all dependencies at once into more manageable, scale-focused sub-problems. This layered decomposition and parallel modeling enhance its ability to capture localized variations, complementing the frequency branch’s focus on global periodicity, leading to a more comprehensive and robust representation.


\subsection{Frequency branch}\label{subsec3.3}
The frequency branch aims to capture global periodic patterns and long-range dependencies by leveraging an autocorrelation mechanism, inspired by Autoformer~\citep{wu2021autoformer}. Unlike time-domain attention, it directly models frequency interactions to extract repeatable structures and phase-aligned dependencies at the sub-series level.

Given the hierarchically sampled frequency components $\tilde{Q}^n,\tilde{K}^n,\tilde{V}^n$, a sub-series similarity is calculated in the frequency domain using the well-known Wiener-Khinchin theorem~\citep{wiener1930generalized}, which links time-domain similarity to frequency-domain operations:
\begin{align*}
    \mathcal{R}_{\mathcal{XX}}(\tau) &=\lim_{L\rightarrow\infty}\frac{1}{L}\sum^{L-1}_{t=1}\mathcal{X}_t\mathcal{X}_{t-\tau}\\
    &=\mathcal{F}^{-1}(\text{Padding}(\tilde{Q}^n\odot(\tilde{K}^n)^*)),
\end{align*}
where $\mathcal{R}_{\mathcal{XX}}(\tau)$ denotes the time-delay similarity at lag $\tau$. Here, $\odot$ represents the element-wise product, and $^*$ denotes complex conjugation. Note that zero-padding is also required before the IFFT operation, like the time branch, to ensure dimension alignment.


After computing the autocorrelation scores $\mathcal{R}_{\tilde{Q}^n \tilde{K}^n}$, which correspond to the attention scores in the standard self-attention mechanism, the top-$k$ most informative lags are selected as follows:
\begin{equation*}
    \tau_1,\dots,\tau_k=\operatorname*{argTop}_{\tau\in\{1,\dots,L\}}k_{\text{lags}}\,(\mathcal{R}_{\tilde{Q}^n \tilde{K}^n}(\tau)).
\end{equation*}

Here, $k_{\text{lags}}$ is the autocorrelation factor and follows the setting in Autoformer~\citep{wu2021autoformer}. The top-$k_{\text{lags}}$ autocorrelation values, $\mathcal{R}_{\tilde{Q}^n \tilde{K}^n}(\tau)$, are then normalized into probabilities using the softmax function for subsequent computations:
\begin{align*}
    \hat{\mathcal{R}}_{\tilde{Q}^n \tilde{K}^n} &(\tau_1),\dots,\hat{\mathcal{R}}_{\tilde{Q}^n \tilde{K}^n}(\tau_{k_{\text{lags}}})\\
    & =\text{Softmax}(\mathcal{R}_{\tilde{Q}^n \tilde{K}^n}(\tau_1),\dots,\mathcal{R}_{\tilde{Q}^n \tilde{K}^n}(\tau_{k_{\text{lags}}})).
\end{align*}

To align the lagged information directly with the current \emph{query}'s time step, the \emph{values} $\tilde{V}^n$ are then cyclically shifted using a \emph{rolling} operator and aggregated based on their respective weights:
\[
\text{AutoCorr}(\tilde{Q}^n,\tilde{K}^n,\tilde{V}^n)=\sum^k_{i=1}Roll(\tilde{V}^n,\tau_i)\hat{\mathcal{R}}_{\tilde{Q}^n\tilde{K}^n}(\tau_i).
\]

This process preserves long-term temporal alignment while allowing efficient modeling of global periodicity. The multi-head variant is defined analogously:
\[
\text{head}_i = \text{AutoCorr}(Q_i,K_i,V_i),
\]
\[
\text{Multi-head}(Q,K,V)= \text{Concat}(\text{head}_1,\dots,\text{head}_h)W^o_f.
\]

\subsection{Periodicity-aware weighting}\label{subsec3.4}
Periodicity is a fundamental characteristic of many real-world time series, typically manifesting as concentrated energy in specific harmonic components~\citep{pukhova2018time}. Specifically, periodic time series normally exhibit concentrated energy at a particular frequency and its harmonics, with sharp spectral peaks in the spectrum indicating stronger periodicity \citep{oppenheim2017signals,pukhova2018time}. Conversely, non-periodic time series generally lack such distinct characteristics, displaying flatter, dispersed energy distributions.

This observation motivates the use of frequency-domain energy concentration as a quantitative proxy for measuring periodicity. Inspired by these insights in the previous work~\citep{wiener1930generalized,ye2024atfnet}, we propose using the harmonic energy ratio as a periodicity measure, which captures the proportion of total spectral energy concentrated in certain harmonics. Although directly computing periodicity is nontrivial, this ratio serves as an interpretable and tractable surrogate that reflects the spectral regularity of the input signal. The following lower bound theorem further convinces of its validity as a periodicity measure.

\begin{theorem}
   Let \( f(t) \) be a continuous time series defined on \( [0, L] \), with its discrete form sampled as $[f(0),\cdots,f(L-1)]$. Assume that \( f(t) \) can be decomposed into\(f(t) = f_p(t) + f_r(t),\)
   where \( f_p(t) \) is the strictly periodic component with period \( \tau \), satisfying $L=m\tau, m\in \mathbb{N}^\ast$, i.e. $f_p(t+\tau) =f_p(t),\:\forall t\in [0,L-\tau]$. And \(f_r(t)\) is the residual component, representing the non-periodic part. Define the following energy metrics:
   \[
   E_p = \sum_{t=0}^{L-1} f_p(t)^2, E_r = \sum_{t=0}^{L-1} f_r(t)^2,
   \]
   \[
   E_f = \sum_{n=0}^{L-1} |F[n]|^2, E_h = \sum_{n \in F_h} |F[n]|^2, F_h=\{k, \dots, nk\},
   \]
   where $E_p$ is energy of periodic part, $E_r$ is energy of residual part, $E_f$ is total spectral energy, and $E_h$ is harmonic energy with a basis frequency $k$.
   
Let \( \lambda = \frac{E_p}{E_r} \) be the energy ratio of the periodic part to the residual part. According to the definition/decomposability of the given time series, we assume that this ratio can be relatively high. Then the lower bound for the ratio of harmonic energy to total energy is given by:
\[
\frac{E_h}{E_f} \geq \frac{\lambda - 2\sqrt{\lambda}}{\lambda - 2\sqrt{\lambda} + 1}.
\]
\end{theorem}
A detailed proof and further discussion are provided in the supplementary material. The result provides a theoretical lower bound for the ratio of harmonic energy to total energy, under the assumption that the time series exhibits sufficiently strong periodicity (i.e., $\lambda >4$). In practice, most of the long-term forecasting datasets satisfy this condition, validating the use of the harmonic energy ratio as a proxy for periodicity. The lower bound ensures that as $\lambda$ increases, the periodicity of the time series strengthens, with greater energy concentration in its harmonic components. Building on this insight, we employ the ratio \( E_h / E_f \) to guide the adaptive weighting of two branches, ensuring a balanced contribution from temporal and frequency features. The specific implementation is detailed in Algorithm \ref{algo}.

Specifically, we compute the harmonic energy ratio from the spectrum of each input sequence, with the basis frequency $k$ automatically identified via spectral peak detection. This method identifies the most prominent low-frequency component with the highest energy, thereby determining the signal's primary periodicity~\citep{palshikar2009simple}. The implementation details are provided in the supplementary material. A higher ratio indicates stronger periodicity and thus assigns greater weight to the frequency branch, while a lower ratio favors the time branch. This adaptive weighting mechanism ensures that Dualformer dynamically allocates attention to local or global features depending on the nature of the input, enhancing its flexibility and robustness across diverse temporal patterns.

\section{Experiments}
In this section, we evaluated the performance of Dualformer through a series of extensive experiments. 
\paragraph{Datasets} We conducted both multivariate and univariate forecasting experiments on eight real-world datasets, including Electricity, Solar energy, Traffic, Weather, and 4 ETT datasets (ETTh1, ETTh2, ETTm1, and ETTm2). Detailed descriptions of the datasets are available in the supplementary material. Following previous work~\citep{wu2021autoformer}, the datasets were split into training, validation, and test sets with a ratio of 6:2:2 for the ETT datasets and 7:1:2 for the rest.

\begin{algorithm}[tb]
    \caption{Periodicity-aware weighting}\label{algo}
    \textbf{Input}: Time series $x$, length $L$, number of harmonics $n$ (set to 3 in experiments)\\
    \textbf{Output}: The weight $w_f,\:w_t$ of the frequency branch and time branch
    \begin{algorithmic}[1] 
        \STATE $F=\text{FFT}(x)$
        \STATE $F_h=\{k,2k,\dots,nk\}$, where basis frequency $k=\text{PeakDetection}(F)$
        \STATE $E_h=\sum_{i\in F_h}|F[i]|^2,\:E_f=\sum_{i=0}^{\lfloor L/2 \rfloor +1}|F[i]|^2$
        \STATE \textbf{return} $w_f= E_h/E_f,\:w_t= 1-w_f$
    \end{algorithmic}
\end{algorithm}

\paragraph{Baselines and experimental setup} We selected several state-of-the-art models as baselines, including TimeMixer~\citep{wang2024timemixerdecomposablemultiscalemixing}, PDF~\citep{dai2024periodicity}, TimesNet~\citep{wu2022timesnet}, FEDformer~\citep{zhou2022fedformer}, FiLM~\citep{zhou2022film}, PatchTST~\citep{nie2022time}, iTransformer\citep{liu2023itransformer}, FreTS~\citep{yi2024frequency}, and DLinear~\citep{zeng2023transformers}. For all models and datasets, the prediction length was set as $T\in \{96, 192, 336, 720\}$, and the lookback window size $L$ was fixed at 96. Both multivariate and univariate predictions were evaluated using widely adopted metrics: Mean Squared Error (MSE) and Mean Absolute Error (MAE). All experiments were repeated three times for robustness. Additional details about the baseline models and experimental setup are provided in the supplementary.

\subsection{Main Results}
The average results across four prediction lengths for multivariate and univariate forecasting are shown in Table \ref{tab:01} and \ref{tab:uni}, respectively, where \textbf{bold}/\underline{underline} values denote the best/second performances. Full results are detailed in the supplementary material. 

Dualformer consistently outperforms baseline models across most datasets, demonstrating its strong capability to handle long-term forecasting. In the multivariate setting, it achieves the top rank in 13 of 16 average cases and 44 of 64 total outcomes across two metrics over eight benchmarks. For the univariate setting, Dualformer also shows competitive performance, consistently ranking among the top methods. However, performance on the Traffic dataset is relatively weaker, likely due to its strong periodicity, which particularly benefits models like PDF that explicitly extract dominant periods. In contrast, Dualformer captures a broader frequency spectrum without isolating specific periodic components, which may limit its advantage on highly seasonal data.

\begin{table*}[ht] \centering 
    \caption{Multivariate long-term time series forecasting results (average) with various prediction length $T\in \{96, 192, 336, 720\}$ and fixed lookback window size $L=96$.  \textbf{Bold} indicates the best and \underline{underline} indicates the second best result, respectively.}\label{tab:01}
    \resizebox{\textwidth}{!}{
    \begin{tabular}{c|cc|cc|cc|cc|cc|cc|cc|cc|cc|cc}
    \toprule
    \textbf{Models} & \multicolumn{2}{c|}{\textbf{Dualformer}} & \multicolumn{2}{c|}{\textbf{TimeMixer}} & \multicolumn{2}{c|}{\textbf{PatchTST}} & \multicolumn{2}{c|}{\textbf{iTransformer}} & \multicolumn{2}{c|}{\textbf{PDF}} & \multicolumn{2}{c|}{\textbf{FEDformer}} & \multicolumn{2}{c|}{\textbf{TimesNet}} & \multicolumn{2}{c|}{\textbf{FiLM}} & \multicolumn{2}{c|}{\textbf{DLinear}} & \multicolumn{2}{c}{\textbf{FreTS}} \\
     & \multicolumn{2}{c|}{(Ours)} & \multicolumn{2}{c|}{(2024)} & \multicolumn{2}{c|}{(2023)} & \multicolumn{2}{c|}{(2024)} & \multicolumn{2}{c|}{(2024)} & \multicolumn{2}{c|}{(2022)} & \multicolumn{2}{c|}{(2023)} & \multicolumn{2}{c|}{(2022)} & \multicolumn{2}{c|}{(2023)} & \multicolumn{2}{c}{(2023)} \\
    \midrule

    Metric & MSE & MAE & MSE & MAE & MSE & MAE & MSE & MAE & MSE & MAE & MSE & MAE & MSE & MAE & MSE & MAE & MSE & MAE & MSE & MAE \\
       \midrule
       \emph{ETTh1} & \textbf{0.407} & \textbf{0.420} & \underline{0.411} & \underline{0.423} &0.418 &0.437 	&0.445 &0.434	& \textbf{0.407} &0.424	&0.433 &0.455	&0.458 &0.450	&0.440 &0.452	&0.439& 0.449	&0.454 &0.447 \\

       \midrule
       \emph{ETTh2} & \underline{0.335} & \textbf{0.377} &	\textbf{0.316} &0.384	&0.343& \underline{0.378} &0.374&0.398	&0.351&0.392	&0.431&0.447	&0.414 &0.427	& 0.359&0.401	&0.458&0.459	&0.383&0.407\\

       \midrule
       \emph{ETTm1} & \textbf{0.345} & \textbf{0.367} & 0.348&0.376	&0.349&0.381&	0.407&0.410	& \underline{0.347} & \underline{0.374}	&0.417&0.440	&0.400& 0.406&	0.361&0.379&	0.355&0.380	&0.414&0.407\\

       \midrule
       \emph{ETTm2} & \textbf{0.245} & \textbf{0.304}	&0.256&0.316	& \underline{0.250} & \underline{0.314} &	0.288&0.332	&0.254&0.315&	0.300&0.348&	0.291 &0.333	&0.254&0.317	&0.281&0.343&	0.286&0.327\\

       \midrule
       \emph{Electricity} & \textbf{0.158} & \textbf{0.249} &	0.165& \underline{0.253} & \underline{0.161} &0.255& \underline{0.161} & 0.257	&0.178&0.270	&0.213&0.326	&0.193&0.295&	0.190&0.283	&0.167&0.264	&0.163&0.259 \\

       \midrule
       \emph{Solar} & \textbf{0.191} & \textbf{0.239} & \underline{0.192} & \underline{0.244} &0.256&0.298&	0.233&0.262	&0.203&0.248	&0.243&0.350	&0.244&0.334	&0.213&0.266	&0.329&0.390	&0.315&0.364 \\

       \midrule
       \emph{Traffic} & 0.406 & \underline{0.268} & \underline{0.393} & 0.271	&0.407&0.282&	0.428&0.282	& \textbf{0.389} & \textbf{0.264} &0.608&0.375	&0.620&0.336&	0.443&0.313&	0.435&0.298&	0.626&0.378  \\

       \midrule
       \emph{Weather} & \textbf{0.220} & \textbf{0.253} &	0.222&0.262	& \underline{0.221} &0.263&	0.258&0.278	&0.222& \underline{0.261}	&0.337&0.377	&0.259&0.287&	0.253&0.284	&0.245&0.297	&0.272&0.291 \\
        
    \bottomrule
    \end{tabular}
    }
\end{table*}

\begin{table*}[ht] \centering 
\caption{Univariate long-term time series forecasting results (average) on ETT datasets with various prediction length $T\in \{96, 192, 336, 720\}$ and fixed lookback window size $L=96$.  \textbf{Bold} indicates the best and \underline{underline} indicates the second best result, respectively.}\label{tab:uni}
    \resizebox{\textwidth}{!}{
    \begin{tabular}{c|c|cc|cc|cc|cc|cc|cc|cc|cc}
     \toprule
     \multicolumn{2}{c|}{Models} & \multicolumn{2}{c|}{Dualformer} & \multicolumn{2}{c|}{Autoformer} & \multicolumn{2}{c|}{FEDformer} & \multicolumn{2}{c|}{FiLM} &
        \multicolumn{2}{c|}{PatchTST} &
        \multicolumn{2}{c|}{iTransformer} &
        \multicolumn{2}{c|}{FreTS} &
        \multicolumn{2}{c}{DLinear} \\
        \midrule
       \multicolumn{2}{c|}{Metric} & MSE & MAE & MSE & MAE &  MSE & MAE &  MSE & MAE &  MSE & MAE &  MSE & MAE&  MSE & MAE&  MSE & MAE \\
       \midrule
       \multicolumn{2}{c|}{\emph{ETTh1}} & \underline{0.073} & \textbf{0.209} & 0.105 & 0.257 & 0.115 & 0.263 & 0.076 & 0.214 & 0.074 & \underline{0.211} & \textbf{0.072} & \textbf{0.209} & 0.102 & 0.248 & 0.098 & 0.240 \\
       \midrule
       \multicolumn{2}{c|}{\emph{ETTh2}} & \textbf{0.173} & \textbf{0.329} & 0.209 & 0.357 & 0.206 & 0.350 & 0.185 & 0.340 & \underline{0.178} & 0.337 & 0.179 & \underline{0.335} & 0.203 & 0.347 & 0.202 & 0.354 \\
       \midrule
       \multicolumn{2}{c|}{\emph{ETTm1}} & \textbf{0.047} & \textbf{0.161} & 0.091 & 0.236 & 0.070 & 0.204 & 0.049 & 0.165 & \underline{0.048} & \underline{0.163} & \underline{0.048} & \underline{0.163} & 0.066 & 0.196 & 0.053 & 0.168 \\
       \midrule
       \multicolumn{2}{c|}{\emph{ETTm2}} & \textbf{0.110} & \textbf{0.247} & 0.157 & 0.305 & 0.130 & 0.273 & 0.115 & 0.255 & 0.113 & 0.251 & \underline{0.111} & \underline{0.246} & 0.118 & 0.259 & 0.113 & 0.249 \\
    \bottomrule
\end{tabular}\label{tab:uni}
}
\end{table*}

\subsection{Ablation Study}\label{sec:4.2}
\begin{figure}[t]
    \centering
    \includegraphics[width=1.0\linewidth]{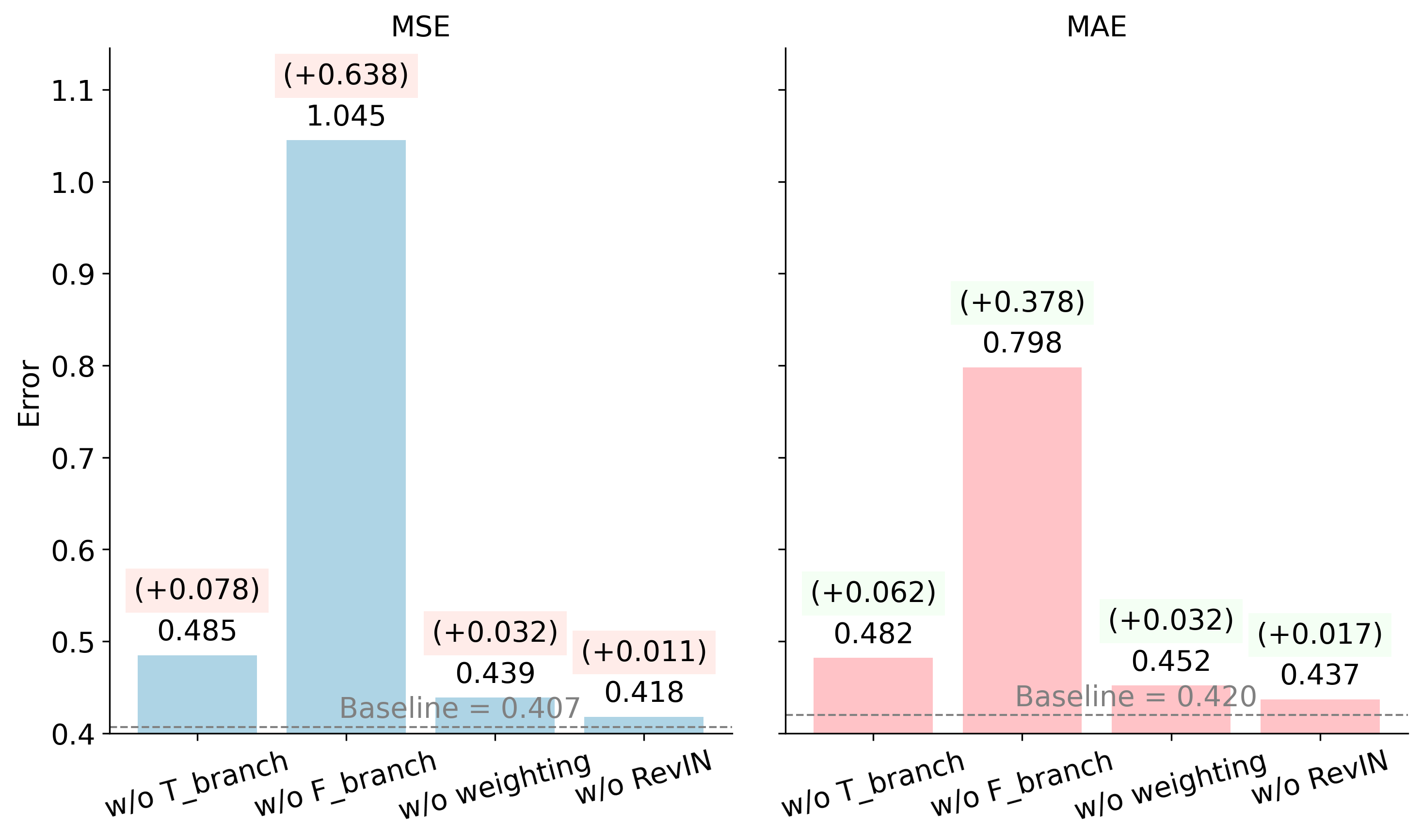}
    \caption{Ablation study results on ETTh1 dataset. Average results of all prediction lengths $T\in\{96,192,336,720\}$ on multivariate forecasting.}
    \label{fig:ablation}
\end{figure}
We further conducted ablation studies to evaluate the contributions of each component in Dualformer. As shown in Figure \ref{fig:ablation}, we compare the full model against four variants: \textbf{w/o T\_branch}: removing the time branch; \textbf{w/o F\_branch}: removing the frequency branch; \textbf{w/o weighting}: replacing the periodicity-aware weighting with uniform averaging; \textbf{w/o RevIN}: removing the RevIN~\citep{kim2021reversible} block.

Removing the frequency branch led to the largest performance drop (+0.638 MSE / +0.378 MAE), highlighting its critical role in modeling global periodicity. The time branch also proved important for capturing local variations. Replacing the periodicity-aware weighting with uniform averaging and removing RevIN caused moderate degradations, confirming the value of input-adaptive fusion and normalization. Full results across all prediction lengths are in the supplementary material.

\subsubsection{Hierarchical frequency sampling evaluation}
To investigate the effectiveness of hierarchical frequency sampling (HFS), we compared it with four fixed or non-adaptive baselines: low-frequency only, high-frequency only, full FFT, and random band selection. Experiments were conducted on two representative datasets with contrasting spectral characteristics: Traffic, dominated by strong periodic patterns with concentrated low-frequency energy; and Weather, which exhibits weak periodicity and contains substantial high-frequency fluctuations and abrupt changes.

\begin{figure}[t]
    \centering
    \includegraphics[width=1.0\linewidth]{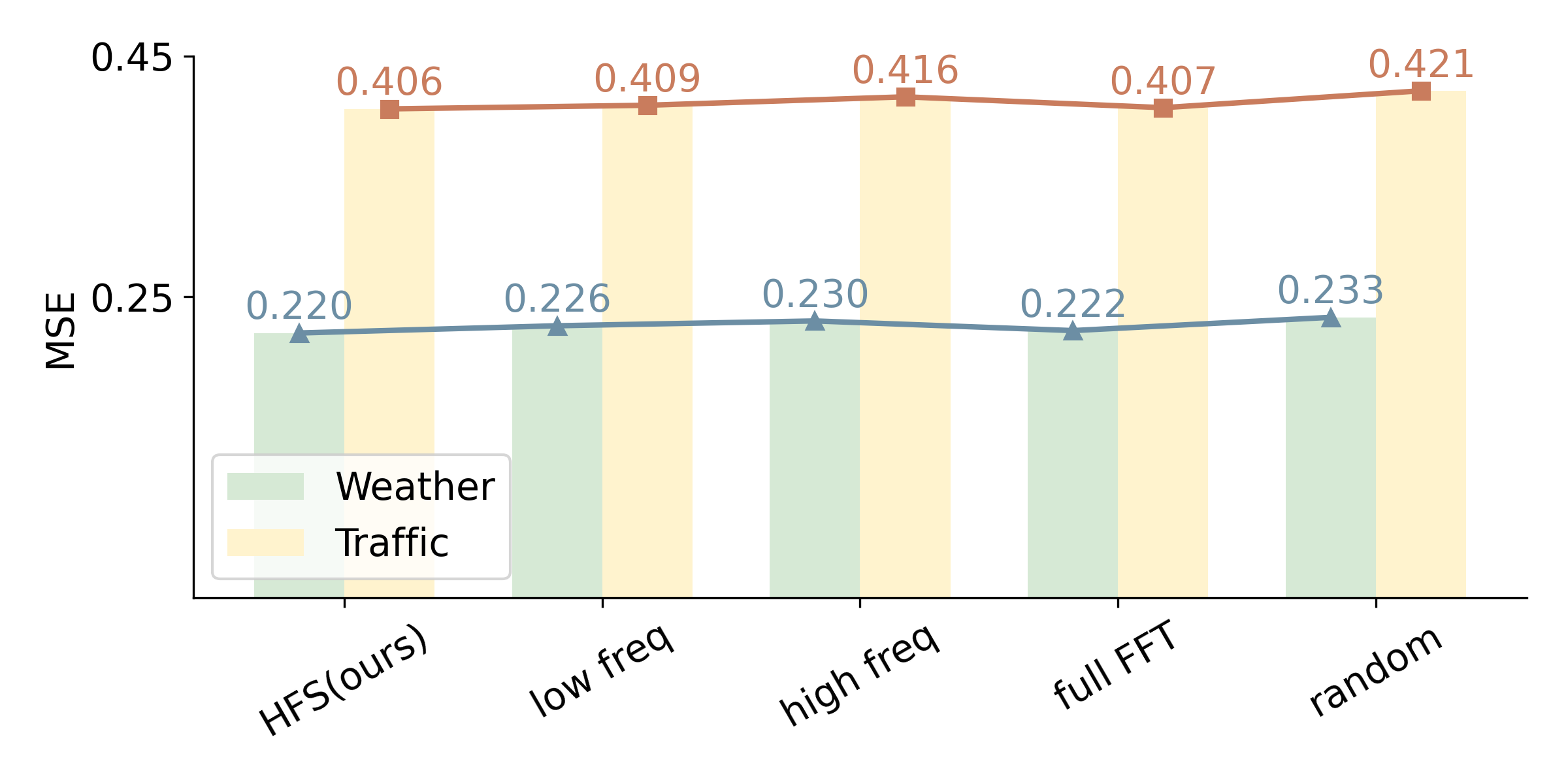}
    \caption{MSE across different frequency selection strategies in two heterogeneous datasets: weather and traffic.}\label{fig:hfs}
\end{figure}

As shown in Figure~\ref{fig:hfs}, HFS consistently achieved the lowest MSE on both datasets, outperforming all baseline strategies. On Traffic, HFS slightly outperforms low-frequency-only and full FFT baselines, indicating its ability to capture dominant periodic patterns while integrating complementary multi-scale information. On Weather, where high-frequency variations are more critical, HFS notably outperforms the high-frequency-only and random strategies, suggesting that static selection is insufficient for datasets with complex or unstable spectral structures.
These results empirically validate that HFS provides a principled way of organizing spectral inputs across model layers. By assigning different frequency bands to different layers, HFS allows the model to gradually process coarse-to-fine structures, thereby improving its capacity to capture both long-term trends and short-term fluctuations. This hierarchical treatment of frequency enhances the model’s representation power and contributes to more accurate long-term forecasting.

\begin{figure}[t]
    \centering
    \includegraphics[width=1.0\linewidth]{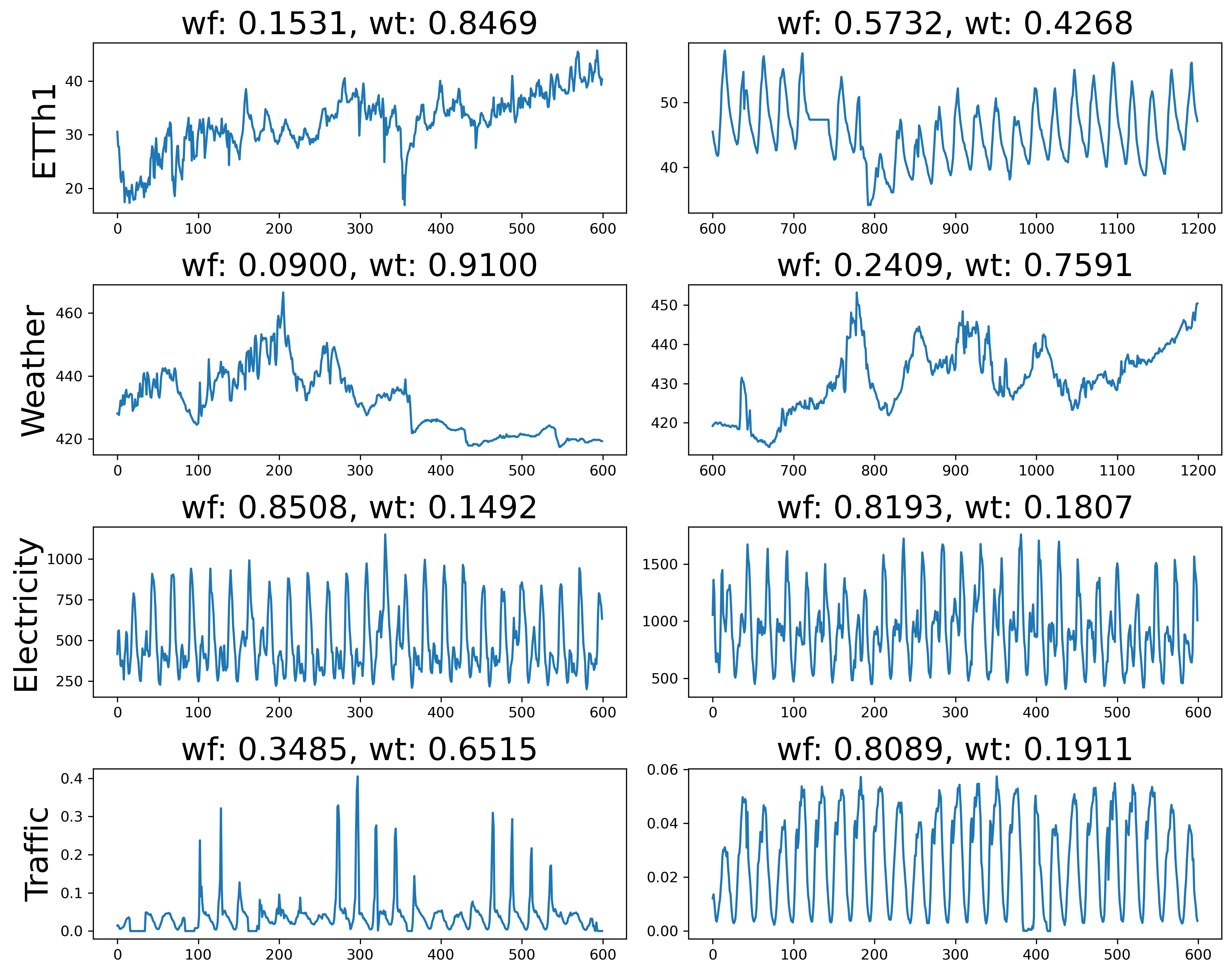}
    \caption{Weight distribution for various time series across four distinct datasets. The examples represent different segments from the ETTh1 and Weather series, and different variables from the Electricity and Traffic datasets.} \label{fig:03}
\end{figure}

\subsubsection{Periodicity-aware weighting evaluation}
To validate the effectiveness of the periodicity-aware weighting mechanism, we analyzed its behavior across four benchmark datasets: ETTh1, Weather, Electricity, and Traffic.

Fig. \ref{fig:03} visualizes the frequency branch weights $w_f$ assigned by Dualformer across eight time series, including different segments or variables from each dataset. For strongly periodic data such as Electricity, the model consistently assigned larger weights to the frequency branch. In contrast, for data with weak or shifting periodicity (e.g., Weather, Traffic, ETTh1), the model adaptively modulated $w_f$ in response to the input, indicating its ability to track temporal variation in spectral structure.

\begin{figure}[t]
  \centering
  \includegraphics[width=0.49\linewidth]{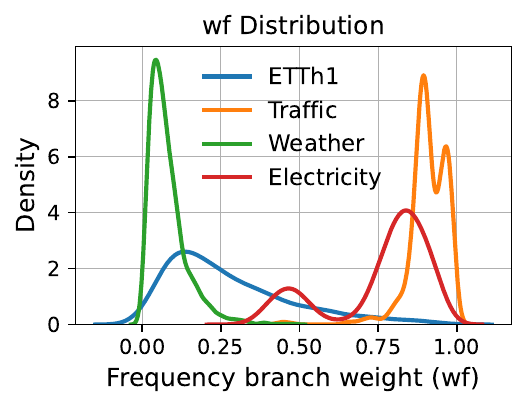}
  \includegraphics[width=0.49\linewidth]{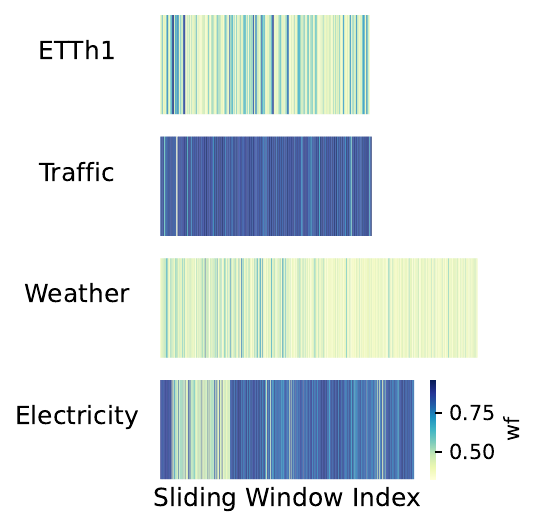}
  \caption{(Left) Distribution of $w_f$ across datasets. (Right) Temporal evolution of $w_f$ under sliding windows.}\label{fig:04}
  \label{fig:wf_case_study}
\end{figure}

Figure \ref{fig:04} (left) shows the global $w_f$ distribution across a 96-sized sliding window. Datasets with stable periodicity (Electricity, Traffic) exhibited distributions skewed toward higher $w_f$, while non-periodic data (Weather) concentrated around lower values. ETTh1 shows a wider spread, reflecting varying degrees of periodicity over time.

Figure \ref{fig:04} (right) further illustrates the temporal evolution of  $w_f$. For example, Electricity maintains a high and stable frequency preference, while Weather exhibits low and fluctuating values, consistent with its non-stationary nature.

These analyses confirm that Dualformer not only captures strong periodic structures by assigning higher weights to the frequency branch but also adapts effectively to weak or unstable periodic patterns by modulating the branch contributions accordingly. This flexibility demonstrates the practical value of the periodicity-aware weighting mechanism, contributing to the model's consistently strong forecasting performance across diverse time series scenarios.

In addition to the above main results, we conducted several supplementary experiments to comprehensively evaluate Dualformer. These include model efficiency (speed and memory usage) test, evaluation on model robustness, an ablation study on the look-back window size, a sensitivity check on key hyperparameters, and performance evaluation on re-normalized data. The results are presented in the supplementary material.

\section{Conclusion}
In this work, we introduced Dualformer, a novel time-frequency dual-domain framework for long-term time series forecasting, which explicitly addresses the low-pass filtering bias of Transformers. Our approach introduced a structured method to model frequency components across network layers and incorporated a dynamic weighting mechanism to balance contributions of time and frequency features based on the input periodicity, enabling adaptive feature prioritization. Extensive experiments on eight benchmarks demonstrate its superior performance over both time- and frequency-domain baselines, highlighting its robustness and adaptability to diverse time series. Future work will explore extensions to high-dimensional multivariate, irregularly sampled, and event-driven time series, as well as broader applications in time series classification and anomaly detection.

\bibliography{aistats26}
\clearpage

\section*{Proof of Theorem 1}
\begin{proof}
Considering the bandlimited property of most time series, here we assume that the energy of those frequency components above the Nyquist frequency($\pi$) is small enough to be neglected, which is also helpful for the following computation. Therefore, for the periodic part $f_p(t)$, the energy $E_p$ is computed by only its first $q=\lceil \pi/2\rceil -1$ Fourier components as follows:
\begin{align}
    E_p &= \sum_{t=0}^{L-1} f_p(t)^2 = \sum_{t=0}^{L-1}\left [ \sum_{m=-q}^q c_me^{2\pi i \frac{m}{\tau}t}\right ]^2 \notag\\
    &= k\sum_{t=0}^{\tau-1}\left [ \sum_{m=-q}^q c_me^{2\pi i \frac{m}{\tau}t}\right ]^2 \notag\\
    &= k\sum_{t=0}^{\tau-1}\sum_{m=-q}^q\sum_{n=-q}^q c_mc_ne^{2\pi i\frac{m+n}{\tau}t} \notag\\
    &= k\sum_{m=-q}^q\sum_{n=-q}^q c_mc_n \sum_{t=0}^{\tau-1}e^{2\pi i\frac{m+n}{\tau}t}
\end{align}
According to the summation of geometric series, we have
\[
\sum_{t=0}^{\tau-1}e^{2\pi i\frac{m+n}{\tau}t}=\frac{1-e^{2\pi i(m+n)}}{1-e^{2\pi i\frac{m+n}{\tau}}}=\begin{cases}
 0 & \text{ if } m+n\ne0  \\
 \tau & \text{ if } m+n=0
\end{cases}
\]
Then the energy of the periodic part is as follows:
\begin{align}
    E_p &= k\sum_{m=-q}^q\sum_{n=-q}^q c_mc_n \sum_{t=0}^{\tau-1}e^{2\pi i\frac{m+n}{\tau}t} \notag\\
    &= k\sum_{m=-q}^q\tau|c_m|^2\notag\\
    &= L\sum_{m=-q}^q|c_m|^2
\end{align}
For the harmonic energy with the basis frequency $k=L/\tau$, we first compute the following Fourier component
\begin{align}
    F[k] &= \sum_{t=0}^{L-1}(f_p(t)+f_r(t))e^{-2\pi i \frac{k}{L}t}\notag\\
    &=\sum_{t=0}^{L-1}\left[ \sum_{m=-q}^q c_me^{2\pi i\frac{m}{\tau}t} + f_r(t)\right]e^{-2\pi i \frac{t}{\tau}}\notag\\
    &= \sum_{t=0}^{L-1} \sum_{m=-q}^q c_me^{2\pi i\frac{m-1}{\tau}t} + \sum_{t=0}^{L-1}f_r(t)e^{-2\pi i \frac{t}{\tau}}\notag\\
    &= c_1L+\sum_{m=-q,m\ne1}^q c_m\sum_{t=0}^{L-1}e^{2\pi i \frac{m-1}{\tau}t}+ \sum_{t=0}^{L-1}f_r(t)e^{-2\pi i \frac{t}{\tau}}\notag\\
    &= c_1L+ \sum_{t=0}^{L-1}f_r(t)e^{-2\pi i \frac{t}{\tau}}
\end{align}      
Similarly, for the harmonics $nk(n=2,\dots,\tau-1)$, we have
\begin{align}
    F[nk] &= \sum_{t=0}^{L-1}(f_p(t)+f_r(t))e^{-2\pi i \frac{nk}{L}t}\notag\\
    &=\sum_{t=0}^{L-1}\left[ \sum_{m=-q}^q c_me^{2\pi i\frac{m}{\tau}t} + f_r(t)\right]e^{-2\pi i \frac{n}{\tau}t}\notag\\
    &= \sum_{t=0}^{L-1} \sum_{m=-q}^q c_me^{2\pi i\frac{m-n}{\tau}t} + \sum_{t=0}^{L-1}f_r(t)e^{-2\pi i \frac{n}{\tau}t}\notag\\
    &= c_nL+\sum_{m=-q,m\ne n}^q c_m\sum_{t=0}^{L-1}e^{2\pi i \frac{m-n}{\tau}t}\notag\\
    & + \sum_{t=0}^{L-1}f_r(t)e^{-2\pi i \frac{n}{\tau}t}\notag\\
    &= c_nL+ \sum_{t=0}^{L-1}f_r(t)e^{-2\pi i \frac{n}{\tau}t}
\end{align}

Then the energy of this harmonic series is computed by
\begin{align}
    E_h &= \sum_{n=0}^{\tau-1}|F[nk]|^2 = \sum_{n=-q}^q|F[nk]|^2 \notag\\
    &= \sum_{n=-q}^q |c_nL+ \sum_{t=0}^{L-1}f_r(t)e^{-2\pi i \frac{n}{\tau}t}|^2 \notag\\
    &= L^2\sum_{n=-q}^q|c_n|^2+2L\sum_{t=0}^{L-1}\sum_{n=-q}^qc_ne^{2\pi i\frac{n}{\tau}t}f_r(t)\notag \\
    &+\sum_{n=-q}^q|\sum_{t=0}^{L-1}f_r(t)e^{-2\pi i\frac{n}{\tau}t}|^2\notag\\
    &\ge L|E_p+2\sum_{t=0}^{L-1}\sum_{n=-q}^qc_ne^{2\pi i\frac{n}{\tau}t}f_r(t)|
\end{align}

As for the total energy $E_f=\sum_{n=0}^{L-1}|F[n]|^2$, according to Parseval's theorem, we have
\begin{align}
    E_f &= \sum_{n=0}^{L-1}|F[n]|^2 = L\sum_{t=0}^{L-1}|f(t)|^2 \notag\\
    &= L\sum_{t=0}^{L-1}\left[ \sum_{m=-q}^qc_me^{2\pi i\frac{m}{\tau}t}+f_r(t)\right]^2\notag\\
    &= L(\sum_{t=0}^{L-1}\left[ \sum_{m=-q}^qc_me^{2\pi i\frac{m}{\tau}t}\right]^2 + \sum_{t=0}^{L-1}f_r^2(t)\notag\\
    & + 2\sum_{t=0}^{L-1}\sum_{m=-q}^qc_me^{2\pi i\frac{m}{\tau}t}f_r(t))\notag\\
    &= L\left[ E_p+E_r+2\sum_{t=0}^{L-1}\sum_{m=-q}^qc_me^{2\pi i\frac{m}{\tau}t}f_r(t)\right]
\end{align}

So it is easy to compute that 
\[
\frac{E_h}{E_f}\ge\frac{E_p+2\sum_{t=0}^{L-1}\sum_{n=-q}^qc_ne^{2\pi i\frac{n}{\tau}t}f_r(t)}{E_p+E_r+2\sum_{t=0}^{L-1}\sum_{n=-q}^qc_ne^{2\pi i\frac{m}{\tau}t}f_r(t)}
\]

The last part satisfies the Cauchy-Schwarz inequality as follows:
\begin{align}
&\left[ \sum_{t=0}^{L-1}\sum_{n=-q}^qc_ne^{2\pi i\frac{m}{\tau}t}f_r(t)\right]^2 \notag\\
&\le \sum_{t=0}^{L-1}\left[\sum_{n=-q}^qc_ne^{2\pi i\frac{m}{\tau}t}\right]^2\cdot\sum_{t=0}^{L-1}f^2_r(t)\notag\\
&\le E_p\cdot E_r
\end{align}
which means that
\[
-\sqrt{E_p\cdot E_r}\le \sum_{t=0}^{L-1}\sum_{n=-q}^qc_ne^{2\pi i\frac{m}{\tau}t}f_r(t) \le \sqrt{E_p\cdot E_r}
\]
Finally, we obtain that
\[
\frac{E_h}{E_f}\ge\frac{E_p-2\sqrt{E_p\cdot E_r}}{E_p+E_r-2\sqrt{E_p\cdot E_r}}=\frac{\lambda-2\sqrt{\lambda}}{\lambda-2\sqrt{\lambda}+1}
\]
\end{proof}

\subsection*{Discussion 1}

Although our proof disregards frequency components above the Nyquist frequency, this simplification is both computationally convenient and theoretically justified. According to the Nyquist-Shannon sampling theorem, frequency components beyond the Nyquist limit cannot be fully reconstructed from the sampling. However, in certain types of data, such as complex physical systems, financial market trends, or physiological signals, high-frequency components are not merely noise; they may encode global features that significantly influence the signal's periodicity or overall trend. 

Given this, it is essential to consider scenarios where high-frequency components are retained and refine the theorem’s assumptions accordingly. 

To accommodate high-frequency contributions, we can extend the Fourier expansion as follows:
\[
E_p'=L[\sum^q_{m=-q}|c_m|^2+\sum_{|m|>q}|c_m|^2]
\]
Note that the same changes occur in the remaining energy computation. Consequently, the energy ratio of the periodic component to the residual component would be redefined as
\[
\lambda'=\frac{E_p+E_{\text{high freq}}}{E_r}
\]
Leading to a revised inequality as
\[
\frac{E_h}{E_f}\ge\frac{\lambda'-2\sqrt{\lambda'}}{\lambda'-2\sqrt{\lambda'}+1}
\]

\subsection*{Discussion 2}
The theorem ensures that when a time series contains a dominant periodic structure (i.e., when $\lambda$ is sufficiently large), the harmonic energy ratio \( E_h / E_f \) is bounded away from zero, offering a degree of theoretical validation. While the bound is not guaranteed for all $\lambda$, especially in weakly periodic settings, it still provides insight into the asymptotic behavior of highly structured signals. In practice, we empirically observe that even mildly periodic sequences exhibit non-negligible harmonic energy concentration (e.g., 
\( E_h / E_f > 0.1 \)), supporting the utility of this ratio as a continuous, data-driven measure of periodicity.

\section*{Experimental Details}
\subsection*{Datasets} Details about the datasets used in this work are given below. Further descriptions are in Table \ref{tab:data}. 

\noindent\textbf{ETT (Electricity Transformer Temperature)}: consists of two hourly datasets (ETTh) and two 15-minute datasets (ETTm), containing loads and oil temperature information of the electricity transformers collected from July 2016 to July 2018.

\noindent\textbf{Electricity}: electricity consumption of 321 customers recorded hourly from 2012 to 2014.

\noindent\textbf{Traffic}: road occupancy rates recorded hourly from highway sensors in San Francisco between 2015 and 2016.

\noindent\textbf{Solar Energy}: solar power production records sampled every 10 minutes from 137 PV plants in Alabama State in 2006.

\noindent\textbf{Weather}: records of 21 meteorological indicators, including air temperature and humidity, collected at 10-minute intervals during 2020.

\begin{table*}[t]
    \caption{Dataset description.}\label{tab:data}
    \centering
    \begin{tabular}{c|c|c|c|c|c}
    \toprule
    Dataset & Source & Frequency & Variate & Length & Time Range  \\
    \midrule
    ETTh1 & Informer\citep{zhou2021informer} & 1h & 7 & 17420 & 2016-2017\\
    \midrule
    ETTh2 & Informer & 1h & 7 & 17420 & 2017-2018\\
    \midrule
    ETTm1 & Informer & 15min & 7 & 69680 & 2016-2017\\
    \midrule
    ETTm2 & Informer & 15min & 7 & 69680 & 2017-2018\\
    \midrule
    Electricity & UCI ML Repository & 1h & 321 & 26304 & 2012-2014\\
    \midrule
    Solar energy & \citep{lai2018modelinglongshorttermtemporal}& 10min & 137 & 52560 & 2006\\
    \midrule
    Traffic & Informer & 1h & 862 & 17544 & 2015-2016\\
    \midrule
    Weather & MPI for Biogeochemistry & 10min &21 & 52696 & 2020\\
    \bottomrule
    \end{tabular}
\end{table*}

\subsection*{Baselines}  
\noindent\textbf{FEDformer}\citep{zhou2022fedformer}(ICML 2022): FEDformer employs seasonal-trend decomposition and a novel attention mechanism applied in the frequency domain to mitigate distribution shift between input and output time series, improving robustness against noise.\\
\noindent\textbf{FiLM}\citep{zhou2022film}(NeurIPS 2022): FiLM (Frequency-Improved Legendre Memory Model) leverages polynomial projections to approximate historical data, applies Fourier projections to eliminate noise, and employs low-rank approximations to enhance computational efficiency.\\
\noindent\textbf{PatchTST}\citep{nie2022time}(ICLR 2023): PatchTST segments time series into patches with predefined window sizes and strides, treating them as input tokens to capture local temporal information. Furthermore, it enforces channel independence, treating each variable in a multivariate time series as a separate univariate sequence while sharing the same embedding and weight structures.\\
\noindent\textbf{iTransformer}\citep{liu2023itransformer}(ICLR 2024): iTransformer reverses the data dimensions, treating each variable as a sequence and each timestep as a feature dimension, enhancing the model's ability to capture inter-variable relationships and temporal trends more effectively.\\
\noindent\textbf{FreTS}\citep{yi2024frequency}(NeurIPS 2023): FreTS transforms time-series data from the time domain to the frequency domain, where an MLP-based architecture is designed to model global dependencies and key patterns. \\
\noindent\textbf{DLinear}\citep{zeng2023transformers}(AAAI 2023): DLinear challenges the effectiveness of Transformers in time-series forecasting by introducing a lightweight architecture that directly enables multi-step prediction. It decomposes time series into a trend series and a remainder series, modeling them separately using two single-layer linear networks.\\
\noindent \textbf{TimeMixer}\citep{wang2024timemixerdecomposablemultiscalemixing}(ICLR 2024): TimeMixer employs a multiscale mixing architecture to address the complex temporal variations in time series forecasting. By utilizing Past-Decomposable-Mixing and Future-Multipredictor-Mixing blocks, TimeMixer leveraged disentangled variations and complementary forecasting capabilities.\\
\noindent \textbf{PDF}\citep{dai2024periodicity}(ICLR 2024): PDF decouples variations of different scales in multi-variate long-term time series based on their periodicity. It then leverages the respective modeling strengths of CNNs and Transformer models to represent these scale-distinct variations. \\
\noindent \textbf{TimesNet}\citep{wu2022timesnet}(ICLR 2023): TimesNet focuses on temporal variation modeling and proposes a representation method intended for incorporating multiple intraperiod- and interperiod-variations. For each selected frequency, the proposed method generates a two-dimensional representation of the original time series and applies a convolutional network to the representation. \\

\subsection*{Implementation details}
All baseline models were implemented based on the Time-Series-Library repository~\citep {wang2024tssurvey}. The experiments were conducted using PyTorch and trained on an NVIDIA A100 40GB GPU.\\
\noindent\textbf{Optimizer}: We employed ADAM with an initial learning rate of $1e^{-4}$, which was adjusted using a cosine decay schedule.\\
\noindent\textbf{Batchsize}: For the Traffic and Electricity datasets, batchsize was set to 32, while for all other datasets, batchsize was set to 256 to ensure efficient training while maintaining stable gradient updates.\\
\noindent\textbf{Early Stopping}: A three-epoch early stopping criterion was used to prevent overfitting.\\
\noindent\textbf{Model Configuration}: The model consists of three encoder layers, designed to effectively capture hierarchical temporal dependencies. The loss function used is L2 loss(Mean Squared Error, MSE), ensuring stable optimization and accurate error minimization.

\subsection*{Related Pseudocodes}
Here, we provide pseudocodes for several core operations in our model. To ensure dimensional alignment after hierarchical frequency sampling, a zero-padding operation is applied in both the time and frequency branches before IFFT transformation, as shown in Algorithm~\ref{pad}. Additionally, for the periodicity-aware weighting mechanism, we employ a simple yet effective peak detection strategy to estimate the basis frequency. This method identifies the frequency component with the maximum amplitude (excluding the DC component) as the dominant periodicity, which is then used to compute harmonic energy. The corresponding procedure is detailed in Algorithm~\ref{peak}.

\begin{algorithm}[H]
\caption{Zero Padding Operation after HFS}\label{pad}
\textbf{Input:} Frequency slice $\tilde{W}\in \mathbb{R}^{B \times F \times C}$, Original frequency length $M$, Start index $p^n$ for $n$-th layer\\
\textbf{Output:} $W_{\text{padded}}\in\mathbb{R}^{B\times M\times C}\text{ padded spectrum}$
\begin{algorithmic}[1]
\STATE Initialize $W_{\text{padded}} \leftarrow \mathbf{0}^{B \times M \times C}$
\STATE Compute end index $q^n \leftarrow p^n + F$
\FOR{$b = 1$ to $B$}
    \FOR{$l = 1$ to $C$}
        \STATE $W_{\text{padded}}[b, p^n : q^n, l] \leftarrow \tilde{W}[b, :, l]$
    \ENDFOR
\ENDFOR
\STATE \textbf{return} $W_{\text{padded}}$
\end{algorithmic}
\end{algorithm}

\begin{algorithm}[H]
\caption{Peak Detection for Periodicity-aware Weighting}
\label{peak}
\textbf{Input:} Magnitude spectrum $|F| \in \mathbb{R}^{B \times M \times C}$\\
\textbf{Output:} Basis frequency index $k \in \mathbb{N}$ for each channel
\begin{algorithmic}[1]
\FOR{$b = 1$ to $B$}
    \FOR{$c = 1$ to $C$}
        \STATE Remove DC component: $|F|_{b,0,c} \leftarrow 0$
        \STATE Normalize: $|F|_{b,:,c} \leftarrow \dfrac{|F|_{b,:,c}}{\sum |F|_{b,:,c|}}$
        \STATE $k_{b,c} \leftarrow \arg\max_{i \ge 1} |F|_{b,i,c}$ \hfill // Find dominant frequency
    \ENDFOR
\ENDFOR
\STATE \textbf{return} $k_{b,c}$ for each $(b,c)$
\end{algorithmic}
\end{algorithm}

\section*{Model Efficiency}
\begin{figure}[t]
    \centering
    \includegraphics[width=1.0\linewidth]{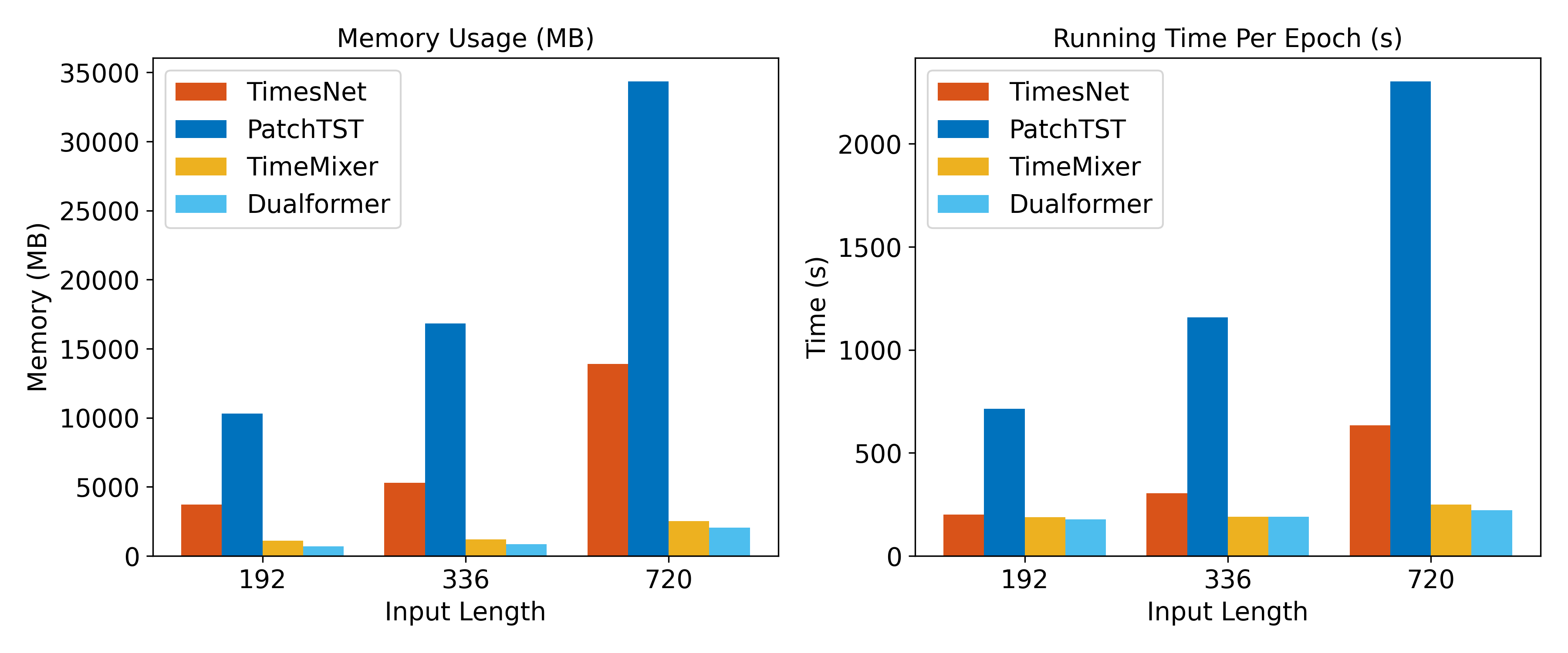}
    \caption{Memory usage and running time per epoch of PatchTST, TimesNet, TimeMixer, and our Dualformer under varying input lengths and fixed prediction length $T=720$ on the Electricity dataset.}
    \label{fig:efficiency}
\end{figure}

We evaluated the efficiency of Dualformer under varying input lengths
$L\in \{192,336,720\}$ on the Electricity dataset with a fixed prediction length of 720. As shown in Figure \ref{fig:efficiency}, Dualformer consistently demonstrates the lowest memory usage (712MB to 2046MB) and fastest running time per epoch (177s to 222s) among all baselines, demonstrating enhanced scalability for longer input sequences.

This efficiency stems from its unique dual-branch architecture and scale-aware processing strategy. Although the model incorporates both time-domain attention ($O(L^2)$) and frequency-domain autocorrelation ($LlogL$), its per-layer theoretical computational complexity remains $O(L^2+\alpha LlogL)$, where $\alpha<1$ denotes the frequency sampling ratio. Crucially, the $O(L^2)$ complexity of the time branch is only nominal. This is because each time-domain layer receives a scale-specific, frequency-filtered input rather than the full, raw signal. Through hierarchical frequency sampling, the model only processes a small subset of spectral components ($\alpha L$ frequency bins),  which are then zero-padded and IFFT-transformed back to the time domain. This provides the time-domain branch with a more refined, less noisy input at a specific scale. This mechanism effectively decomposes global modeling into localized, scale-aware sub-tasks, which significantly reduces the effective computation required by the time-domain attention in practice. By minimizing redundant information and attention interference, it enables efficient learning while effectively lowering the constant factor associated with the $L^2$ term, despite retaining the nominal quadratic complexity.

Furthermore, the time and frequency branches are designed to execute in parallel. This parallelization leads to improved GPU utilization, thereby reducing overall runtime. While parallel execution can often increase instantaneous memory peaks, Dualformer effectively manages memory overhead through optimized memory management and activation re-computation strategies (e.g., recomputing certain intermediate activations during gradient calculation instead of storing them throughout\citep{korthikanti2022reducingactivationrecomputationlarge}), resulting in overall memory efficiency that remains significantly superior to other baselines.


\begin{figure}[t]
    \centering
    \includegraphics[width=1.0\linewidth]{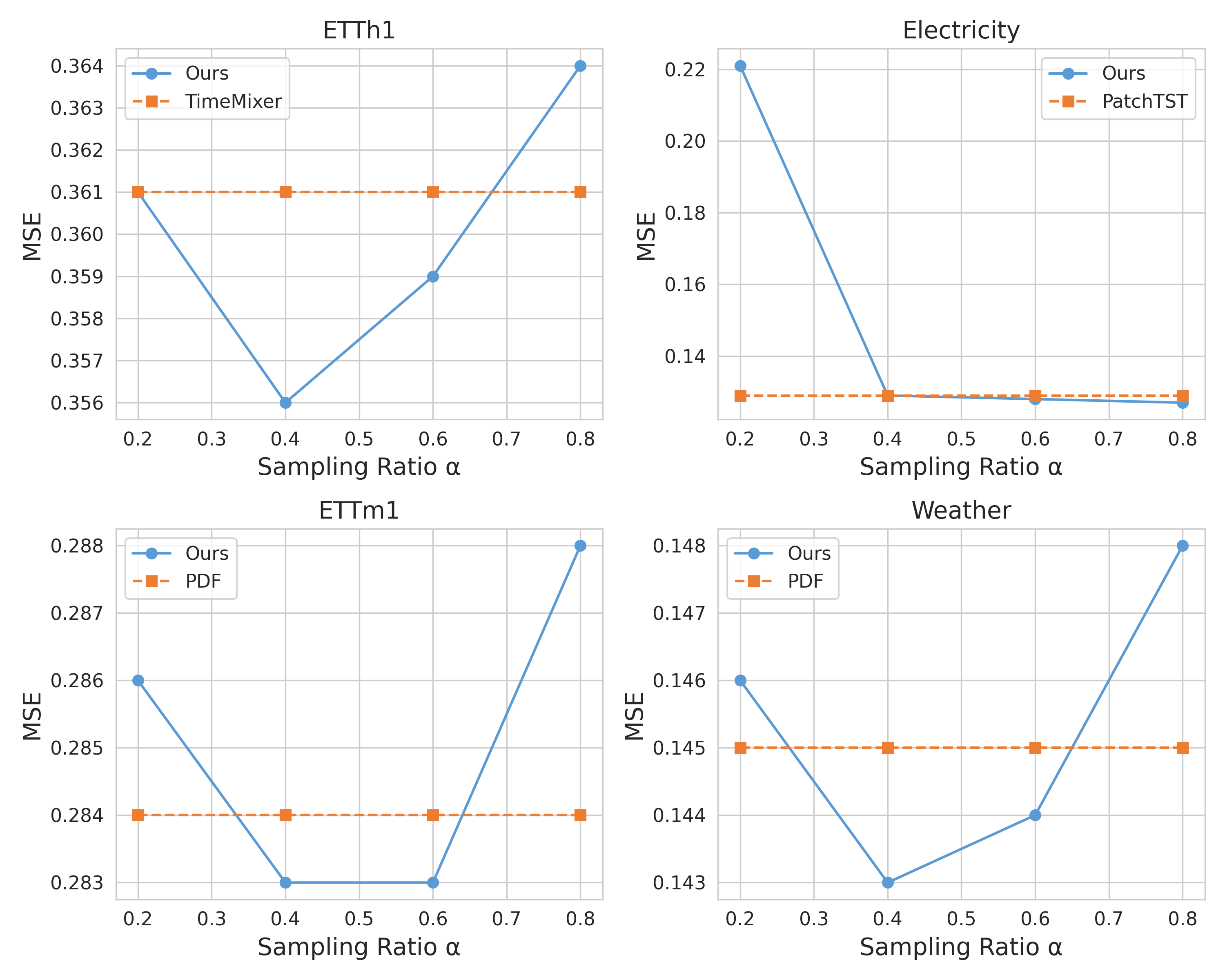}
    \caption{MSE comparison w.r.t different sampling ratios on four datasets. The orange lines indicate baselines with second-best performance.}
    \label{fig:sampling_ratio}
\end{figure}

\begin{figure}[t]
    \centering
    \includegraphics[width=1.0\linewidth]{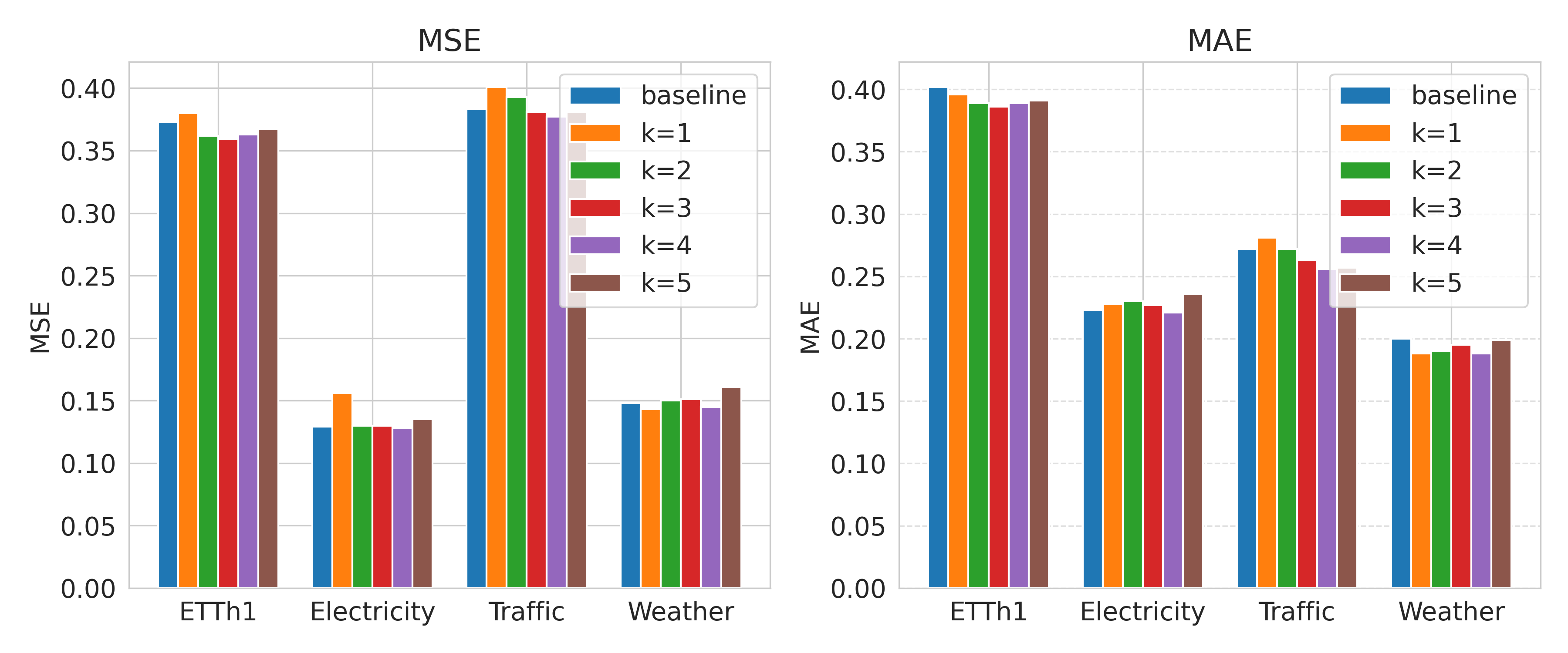}
    \caption{Performance comparison w.r.t different top-$k$ time delay aggregation on fourt datasets.}
    \label{fig:auto}
\end{figure}

\section*{Hyperparameter Sensitivity}

\subsection*{Sampling ratio $\alpha$}

We further conducted a sensitivity analysis on the sampling ratio $\alpha$, which controls the range of frequency components selected at each layer. This analysis was under an input-96-predict-96 setting on the ETTh1, Electricity, ETTm1, and Weather datasets. As shown in Fig.~\ref{fig:sampling_ratio}, the optimal value of $\alpha$ varies across datasets. For instance, lower sampling ratios (e.g., $\alpha=0.4$) yield better performance on ETT and Weather datasets, likely due to the importance of high-frequency variations. In contrast, the Electricity dataset shows less sensitivity to $\alpha$, possibly due to its dominant low-frequency patterns. Notably, across all datasets, our model with a well-tuned $\alpha$ surpasses the performance of strong baseline models, highlighting the importance of tuning this hyperparameter to the specific characteristics of the time-series data to achieve optimal results.

\subsection*{Autocorrelation factor $k_\text{lags}$}
We also conducted a sensitivity analysis on the autocorrelation factor $k_\text{lags}$ to investigate its impact on prediction performance. As shown in Figure \ref{fig:auto}, we compared the baseline model (PatchTST) with different values of $k_\text{lags}$ ranging from 1 to 5 under the input-96-predict-96 setting. The results indicate that moderate values of $k_\text{lags}$ generally lead to improved accuracy. For instance, $k_\text{lags}=3$ yields the best performance on ETTh1, while $k_\text{lags}=4$ achieves the lowest error on both Electricity and Traffic datasets. In contrast, the Weather dataset benefits most from a smaller $k_\text{lags}=1$, suggesting that overly aggregating delayed dependencies may introduce noise for data with weak temporal patterns. These findings highlight the importance of carefully tuning $k_\text{lags}$ to balance dependency modeling and noise suppression.

\subsection*{Number of harmonics $n$}
\begin{figure*}[t]
    \centering
    \includegraphics[width=.9\textwidth]{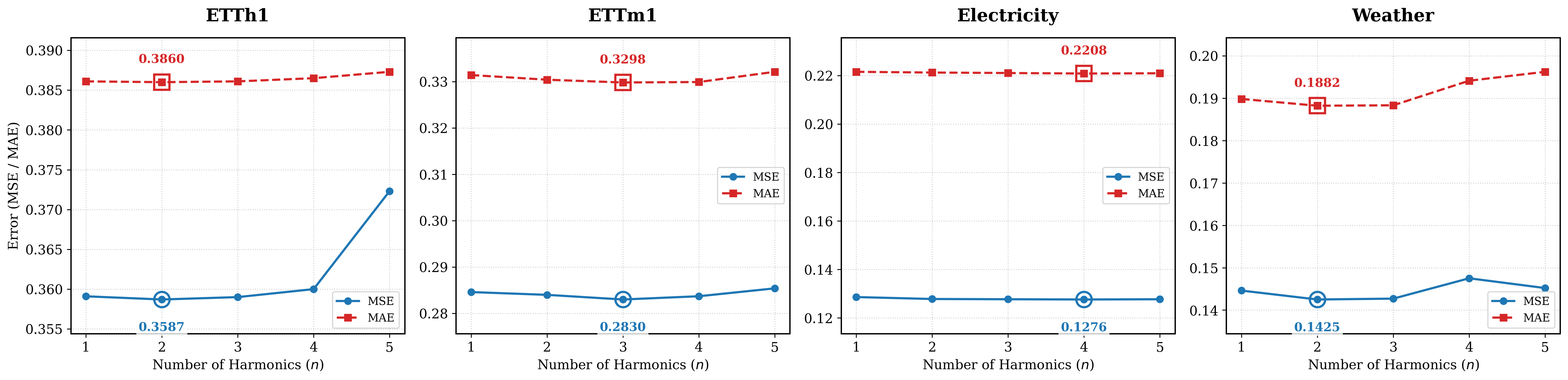}
    \caption{Performance comparison w.r.t different number of harmonics $n$ on four datasets.}
    \label{fig:n}
\end{figure*}

We similarly conducted a sensitivity analysis on the number of harmonics $n$ utilized in the periodicity-aware weighting algorithm. We varied $n$ from 1 to 5 under the input-96-predict-96 setting across the ETTh1, ETTm1, Electricity, and Weather datasets. As shown in Figure \ref{fig:n}, while the exact optimal value exhibits slight variations across datasets (e.g., $n=4$ for Electricity and $n=2$ for Weather), setting $n=3$ generally achieves the best or near-optimal forecasting accuracy. Specifically, we observe that relying on a single harmonic ($n=1$) often yields suboptimal results, whereas increasing $n$ beyond 3 offers marginal gains and, in some cases (such as ETTh1), leads to performance degradation likely caused by noise from higher-order harmonics.

\begin{table}[t]
    \centering
    \resizebox{.5\textwidth}{!}{
    \begin{tabular}{cc|cccc}
        \toprule
        \textbf{Dataset} & \textbf{Metrics} & DLinear & iTransformer & Dualformer & PathTST \\
        \midrule
        \multirow{3}{*}{\emph{ETTh1}}
         &MAE & 1.76 & 1.71 & \underline{1.60} & \textbf{1.58} \\
         
         &RMSE & 3.12 & 3.15 & \underline{3.08} & \textbf{3.06}  \\
         
         &WAPE & 37.61\% & 36.89\% & \underline{34.49\%} & \textbf{34.06\%}   \\

         \midrule
        \multirow{3}{*}{\emph{ETTm1}}
         &MAE & 1.54 & 1.53 & \textbf{1.37} & \underline{1.39} \\
         
         &RMSE & 2.92 & \underline{2.89} & 3.01 & \textbf{2.78}  \\
         
         &WAPE & 33.36\% & 33.03\% & \textbf{29.60\%} & \underline{30.02\%}   \\

         \midrule
        \multirow{3}{*}{\emph{Electricity}}
         &MAE & 281.63 & 295.98  & \textbf{253.57} & \underline{256.60} \\
         
         &RMSE & 2936.83 & \textbf{2761.75} & 2896.04  & \underline{2883.63}  \\
         
         &WAPE & 10.56\% & 11.91\% & \underline{9.51\%} & \textbf{9.38\%}   \\

         \midrule
        \multirow{3}{*}{\emph{Weather}}
         &MAE & 12.29 & 15.61 & \textbf{10.85} & \underline{12.08} \\
         
         &RMSE & 44.57 & \textbf{40.74} & \underline{41.70} & 43.22  \\
         
         &WAPE & 7.29\% & \underline{6.75\%} & \textbf{6.49\%} & 7.32\%   \\

        \bottomrule
    \end{tabular}
    }
    \caption{Average forecasting results on re-normalized data. \textbf{Bold} indicates the best performance, while \underline{underline} indicates the second best.}\label{tab:02}
\end{table}

\begin{figure}[t]
    \centering
    \includegraphics[width=0.85\linewidth]{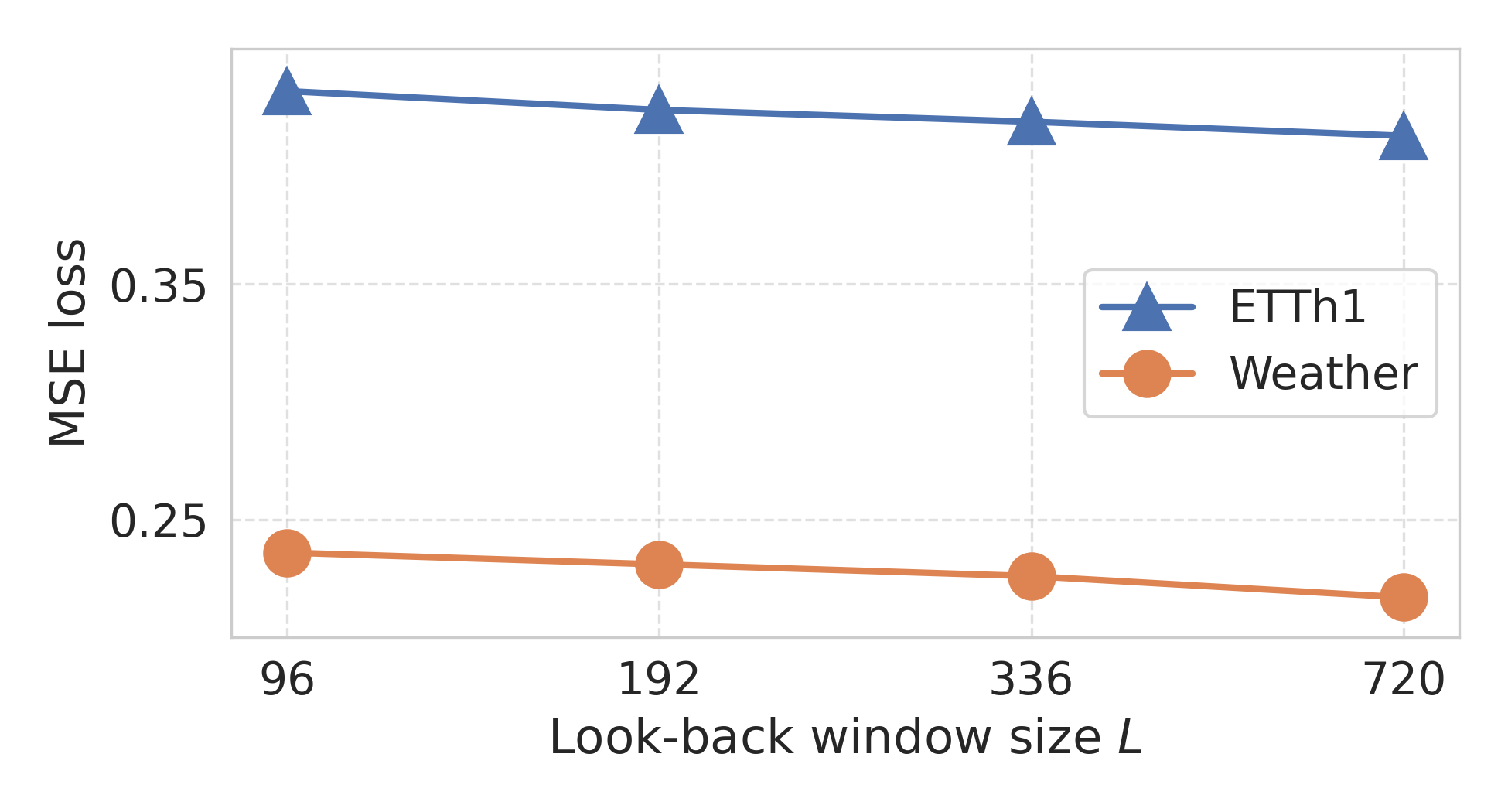}
    \caption{The forecasting performance on ETTh1 and Weather datasets, across 4 different look-back window sizes. The prediction length was fixed as $T=336$.}
    \label{fig:window}
\end{figure}

\begin{table*}[t]
    \resizebox{\textwidth}{!}{
    \begin{tabular}{c|cc|cc|cc|cc|cc}
        \toprule
         & \multicolumn{2}{c|}{Original} & \multicolumn{2}{c|}{w/o T\_branch} & \multicolumn{2}{c|}{w/o F\_branch} & \multicolumn{2}{c}{w/o weighting}& \multicolumn{2}{c}{w/o RevIN} \\
        \midrule
        & MSE & MAE & MSE & MAE & MSE & MAE & MSE & MAE &  MSE & MAE\\
        \midrule
         96 & 0.359 & 0.386 & 0.449 & 0.455 & 0.916 & 0.741  & 0.395 & 0.419 & 0.377 & 0.396 \\
         
         192 & 0.398 & 0.412 & 0.468 & 0.465 & 0.975 & 0.765 & 0.424 & 0.445 &0.409 & 0.437\\
         
         336 & 0.432 & 0.431 & 0.506 & 0.494 & 1.099 & 0.819 & 0.446 & 0.466 &0.440 & 0.449 \\
         
         720 & 0.438 & 0.453 & 0.516 & 0.513 & 1.191 & 0.865 & 0.491 & 0.468  &0.446 & 0.466\\
         
        \midrule
        \emph{Avg.} & 0.407 & 0.420 & 0.485(+0.078)& 0.482(+0.062)& 1.045(+0.638) & 0.798(+0.378)& 0.440(+0.033) & 0.449(+0.028) & 0.439(+0.032) & 0.452(+0.031) \\
        \bottomrule
    \end{tabular} 
    }
    \caption{Ablation study results on ETTh1 dataset.}\label{tab:03}
\end{table*}

\begin{table*}[t] \centering 
    \resizebox{\textwidth}{!}{
    \begin{tabular}{c|c|cc|cc|cc|cc}
        \toprule
        \multicolumn{2}{c|}{Models} &\multicolumn{2}{c|}{\textbf{Dualformer}} & \multicolumn{2}{c|}{\textbf{TimeMixer}} & \multicolumn{2}{c|}{\textbf{PatchTST}} & \multicolumn{2}{c}{\textbf{iTransformer}} \\
        \midrule
       \multicolumn{2}{c|}{Metric} & MSE & MAE & MSE & MAE &  MSE & MAE &  MSE & MAE \\
        \midrule
        \multirow{4}{*}{\rotatebox{90}{\emph{ETTh1}}}
         &96 & \textbf{0.359 $\pm$ 0.0014} & \textbf{0.386 $\pm$ 0.0003}	& 0.361 $\pm$ 0.0003 & 0.390 $\pm$ 0.0002 & 0.373 $\pm$ 0.0009 & 0.402 $\pm$ 0.0008 & 0.381  $\pm$ 0.0021 & 0.395 $\pm$ 0.0024\\
         
         &192 & \textbf{0.398 $\pm$ 0.0015}& \textbf{0.412 $\pm$ 0.0010} &	0.409 $\pm$ 0.0005 & 0.414 $\pm$ 0.0008	&0.411 $\pm$ 0.0014 &0.428 $\pm$ 0.0015	&0.437 $\pm$ 0.0020 &0.424 $\pm$ 0.0022\\
         
         &336 &0.432 $\pm$ 0.0011& 0.431 $\pm$ 0.0007 &	\textbf{0.430 $\pm$ 0.0011}& \textbf{0.429 $\pm$ 0.0013}	&0.432$\pm$ 0.0059 &0.445 $\pm$ 0.0048 &	0.479 $\pm$ 0.0040 &0.446 $\pm$ 0.0037\\
         
         &720 & \textbf{0.438 $\pm$ 0.0016} & \textbf{0.453 $\pm$ 0.0016} & 0.445 $\pm$ 0.0034 &0.460 $\pm$ 0.0033	&0.455 $\pm$ 0.0049 &0.473 $\pm$ 0.0038&	0.481 $\pm$ 0.0055 &0.470 $\pm$ 0.0039\\
         
        \midrule
         
        \multirow{4}{*}{\rotatebox{90}{\emph{ETTm1}}}
         &96 & \textbf{0.283 $\pm$ 0.0010} & \textbf{0.330 $\pm$ 0.0006} &0.291 $\pm$ 0.0005 &0.340 $\pm$ 0.0004	&0.288 $\pm$ 0.0006& 0.341 $\pm$ 0.0010 &	0.334 $\pm$ 0.0004& 0.368 $\pm$ 0.0004\\
         
         &192 & \textbf{0.326 $\pm$ 0.0007} & \textbf{0.355 $\pm$ 0.0005} & 0.327 $\pm$ 0.0001 & 0.365 $\pm$ 0.0004 & 0.332 $\pm$ 0.0009 &0.370 $\pm$ 0.0007& 0.377 $\pm$ 0.0014 &0.391 $\pm$ 0.0006\\
         
         &336 & \textbf{0.359 $\pm$ 0.0009} & \textbf{0.377 $\pm$ 0.0005}	&0.360 $\pm$ 0.0022& 0.381 $\pm$ 0.0018	&0.362 $\pm$ 0.0018& 0.392 $\pm$ 0.0002 &	0.426 $\pm$  0.0008&0.420 $\pm$ 0.0004  \\
         
         &720 & \textbf{0.412 $\pm$ 0.0013} & \textbf{0.408 $\pm$ 0.0005} & 0.415 $\pm$ 0.0001 & 0.417 $\pm$ 0.0003	&0.416 $\pm$ 0.0002& 0.419 $\pm$ 0.0003 &0.491 $\pm$ 0.0009 &0.459  $\pm$ 0.0007\\

         \midrule
         
        \multirow{4}{*}{\rotatebox{90}{\emph{Electricity}}}
         &96 & \textbf{0.128 $\pm$ 0.0001} & \textbf{0.221$\pm$ 0.0001}	& 0.135 $\pm$ 0.0003& 0.222 $\pm$ 0.0004	& 0.129 $\pm$ 0.0005 & 0.223 $\pm$ 0.0005	& 0.131 $\pm$ 0.0000 & 0.227$\pm$ 0.0002 \\
         
         &192 &  \textbf{0.145 $\pm$0.0002} & 0.237 $\pm$ 0.0001 & 0.147 $\pm$ 0.0006 & \textbf{0.235$\pm$0.0007} & 0.149 $\pm$ 0.0006& 0.243 $\pm$ 0.0014 & 0.148 $\pm$ 0.0002& 0.244 $\pm$ 0.0003 \\
         
         &336 &  \textbf{0.160 $\pm$ 0.0004} & 0.253 $\pm$ 0.0004 & 0.164 $\pm$ 0.0004 & \textbf{0.245 $\pm$ 0.0003} & 0.165 $\pm$ 0.0010 &0.260 $\pm$ 0.0012& 0.164 $\pm$ 0.0001 &0.262	$\pm$ 0.0003\\
         
         &720 &  \textbf{0.197 $\pm$ 0.0010} & \textbf{0.285 $\pm$ 0.0006} &	0.212 $\pm$ 0.0005 & 0.310 $\pm$ 0.0004	& 0.200 $\pm$ 0.0027 & 0.292 $\pm$ 0.0020 & 0.200 $\pm$ 0.0003 &0.295 $\pm$ 0.0002	\\

         \midrule
         
        \multirow{4}{*}{\rotatebox{90}{\emph{Traffic}}}
         &96 & 0.377 $\pm$ 0.0015 & \textbf{0.256 $\pm$ 0.0004}	& \textbf{0.366 $\pm$ 0.0007} & 0.259	$\pm$ 0.0008 & 0.383 $\pm$ 0.0182 &0.272 $\pm$ 0.0174	& 0.395 $\pm$ 0.0012& 0.268 $\pm$ 0.0001 \\
         
         &192 & 0.392 $\pm$ 0.0001 & 0.263 $\pm$ 0.0001 &	0.381 $\pm$ 0.0005 & 0.265 $\pm$ 0.0003 & \textbf{0.380 $\pm$ 0.0011} & \textbf{0.259 $\pm$ 0.0007} & 0.417 $\pm$ 0.0001& 0.276 $\pm$ 0.0002\\
         
         &336 &   0.406 $\pm$ 0.0008& \textbf{0.266 $\pm$ 0.0007} & \textbf{0.397 $\pm$ 0.0008} & 0.269 $\pm$ 0.0008 & 0.410 $\pm$ 0.0129 &0.285 $\pm$ 0.0147	&0.433 $\pm$ 0.0006& 0.283 $\pm$ 0.0008 \\
         
         &720 & 0.447 $\pm$ 0.0020& \textbf{0.287 $\pm$ 0.0008} & \textbf{0.429 $\pm$ 0.0016} &0.292 $\pm$ 0.0019	&0.454 $\pm$ 0.0018&0.313 $\pm$ 0.0014&	0.467 $\pm$ 0.0009& 0.302$\pm$0.0002\\

        \bottomrule
    \end{tabular} }
    \caption{Multivariate long-term forecasting results with error bars (Mean ± STD). All experiments were repeated 3 times. \textbf{Bold} indicates the best.}\label{tab:error_bar}
\end{table*}

\section*{Supplementary of Main Results}

\subsection*{Ablation study}
We conducted several ablation studies to demonstrate the contribution of each model component on the ETTh1 dataset in section 4.2. The full results are available in Table \ref{tab:03}.

\begin{table*}[t] \centering
    \resizebox{\textwidth}{!}{
    \begin{tabular}{c|c|cc|cc|cc|cc|cc|cc|cc|cc|cc|cc}
        \toprule
        \multicolumn{2}{c|}{Models} &\multicolumn{2}{c|}{\textbf{Dualformer}} & \multicolumn{2}{c|}{\textbf{TimeMixer}} & \multicolumn{2}{c|}{\textbf{PatchTST}} & \multicolumn{2}{c|}{\textbf{iTransformer}} & \multicolumn{2}{c|}{\textbf{PDF}} & \multicolumn{2}{c|}{\textbf{FEDformer}} & \multicolumn{2}{c|}{\textbf{TimesNet}} & \multicolumn{2}{c|}{\textbf{FiLM}} & \multicolumn{2}{c|}{\textbf{DLinear}} & \multicolumn{2}{c}{\textbf{FreTS}} \\
     \multicolumn{2}{c|}{} &\multicolumn{2}{c|}{(Ours)} & \multicolumn{2}{c|}{(2024)} & \multicolumn{2}{c|}{(2023)} & \multicolumn{2}{c|}{(2024)} & \multicolumn{2}{c|}{(2024)} & \multicolumn{2}{c|}{(2022)} & \multicolumn{2}{c|}{(2023)} & \multicolumn{2}{c|}{(2022)} & \multicolumn{2}{c|}{(2023)} & \multicolumn{2}{c}{(2023)} \\
        \midrule
       \multicolumn{2}{c|}{Metric} & MSE & MAE & MSE & MAE &  MSE & MAE &  MSE & MAE &  MSE & MAE &MSE & MAE &MSE & MAE &MSE & MAE &MSE & MAE &MSE & MAE \\
        \midrule
        \multirow{4}{*}{\rotatebox{90}{\emph{ETTh1}}}
         &96 & \textbf{0.359} & \textbf{0.386}	&\underline{0.361}& \underline{0.390}&	0.373 & 0.402&	0.381& 0.395 	&0.369 &0.397	& 0.376& 0.416&	0.384& 0.402&	0.377 &0.401&	0.375& 0.396&	0.386& 0.405\\
         
         &192 & \textbf{0.398}& \textbf{0.412} &	0.409 &\underline{0.414}	&0.411 &0.428	&0.437 &0.424& 	\underline{0.401} & 0.416	&0.422 &0.445	&0.436 &0.429 & 0.419& 0.428	&0.428& 0.437	&0.441& 0.436\\
         
         &336 &0.432& 0.431 &	\underline{0.430}& \underline{0.429}	&0.432 &0.445 &	0.479 &0.446	&\textbf{0.418} & \textbf{0.427} &	0.452 &0.463&	0.491& 0.469&	0.466& 0.466 &	0.448 &0.449	&0.487& 0.458 \\
         
         &720 & \textbf{0.438} & \textbf{0.453} & 0.445 &0.460	&0.455 &0.473&	0.481 &0.470	& \underline{0.439} & \underline{0.454}	&0.483 & 0.496& 	0.521& 0.500&	0.499& 0.512 &	0.505& 0.514 & 0.503 &0.491 \\
         
        \midrule
        
        \multirow{4}{*}{\rotatebox{90}{\emph{ETTh2}}}
         &96 & \textbf{0.268} & \textbf{0.329} & \underline{0.271} & \underline{0.330} &	0.275& 0.338&	0.288& 0.338&	0.274& 0.337 &0.343& 0.385 & 0.340 &0.374	&0.280& 0.343 & 0.296& 0.360	&0.297& 0.349 \\
         
         &192 & \underline{0.332} & \textbf{0.370} &	\textbf{0.317} & 0.402&	0.338 & \underline{0.380} &	0.374 &0.390&	0.351& 0.383&	0.429 &0.438 &	0.402& 0.414 & 0.345& 0.390 &	0.391& 0.423 & 0.380& 0.400 \\
         
         &336 & \underline{0.359} & \textbf{0.393} &	\textbf{0.332} & \underline{0.396} & 0.366 &0.403	& 0.415 &0.426	&0.375 &0.411&	0.489 &0.485 & 0.452 & 0.452 & 0.372 &0.415 & 0.445& 0.460 & 0.428& 0.432\\
         
         &720 & \underline{0.381} & \underline{0.417} & \textbf{0.342} & \textbf{0.408} &	0.391& 0.431 & 0.420 &0.440&	0.402 & 0.438 & 0.463& 0.481 &	0.462& 0.468 & 0.438& 0.455 & 0.700 &0.592 & 0.427 &0.445  \\

         \midrule
         
        \multirow{4}{*}{\rotatebox{90}{\emph{ETTm1}}}
         &96 & \textbf{0.283} & \textbf{0.330} &0.291 &0.340	&0.288& 0.341 &	0.334& 0.368	& \underline{0.284} & \underline{0.339} &	0.356& 0.406 & 0.338& 0.375& 0.303& 0.345 & 0.301 &0.344 & 0.355& 0.376 \\
         
         &192 & \underline{0.326} & \textbf{0.355} & 0.327 & 0.365 & 0.332 &0.370 & 0.377 &0.391	& \textbf{0.322} & \underline{0.362}	&0.391 &0.424 &	0.327& 0.387 & 0.342 &0.369 & 0.336 &0.366 & 0.391 &0.392  \\
         
         &336 & \textbf{0.359} & \textbf{0.377}	&0.360& 0.381	&0.362& 0.392 &	0.426 &0.420 & \underline{0.360} & \underline{0.380} & 0.441 &0.453	&0.410 &0.411 & 0.371 &0.387 & 0.371& 0.387 & 0.424 &0.415  \\
         
         &720 & \textbf{0.412} & \textbf{0.408} & \underline{0.415}& 0.417	&0.416& 0.419 &0.491 &0.459 & 0.421& \underline{0.416} & 0.482& 0.476& 0.478& 0.450 & 0.430& 0.416 & 0.426 &0.422 & 0.487& 0.450\\

         \midrule
         
        \multirow{4}{*}{\rotatebox{90}{\emph{ETTm2}}}
         &96 & \textbf{0.157} & \textbf{0.244} &0.164& \underline{0.254} &\underline{0.162}& \underline{0.254} &	0.180 &0.264 & 0.167& 0.258 & 0.189& 0.281& 0.187& 0.267 & 0.167 & 0.257 & 0.170& 0.264 & 0.182& 0.265\\
         
         &192 & \textbf{0.213} & \textbf{0.282} & 0.223 & 0.295 & \underline{0.217} & \underline{0.293} & 0.250 &0.309 & 0.221 & 0.294	& 0.257 &0.324	& 0.249 & 0.309	& 0.219 & 0.293 & 0.233 &0.311 & 0.246 &0.304\\
         
         &336 & \textbf{0.264} & \textbf{0.318} & 0.279 & 0.330 & \underline{0.267} & \underline{0.326} & 0.311 &0.348	& 0.270 & 0.327	& 0.325 &0.364	& 0.321 & 0.351	& 0.273 &0.331	& 0.300& 0.358	& 0.307 &0.342 \\
         
         &720 & \textbf{0.345} & \textbf{0.371} & 0.359 &0.383 & \underline{0.353} & \underline{0.382} & 0.412 &0.407	& 0.356 & \underline{0.382} & 0.429& 0.424 & 0.408 & 0.403	& 0.356& 0.387 & 0.422 &0.439 & 0.407& 0.398 \\

         \midrule
         
        \multirow{4}{*}{\rotatebox{90}{\emph{Electricity}}}
         &96 & \textbf{0.128} & \textbf{0.221}	& 0.135 & \underline{0.222}	& \underline{0.129} & 0.223	& 0.131& 0.227 & 0.148 & 0.240	& 0.189 &0.305 & 0.168 & 0.272	& 0.154& 0.248	& 0.140 &0.237 & 0.133 &0.229\\
         
         &192 &  \textbf{0.145} & \underline{0.237} & \underline{0.147} & \textbf{0.235} & 0.149& 0.243 & 0.148& 0.244 &	0.162 & 0.253 & 0.205& 0.320 &	0.184 & 0.289 & 0.167& 0.260 &	0.154 &0.250 & 0.152 &0.248\\
         
         &336 &  \textbf{0.160} & \underline{0.253} & \underline{0.164} & \textbf{0.245} & 0.165 &0.260 & 0.164 &0.262	& 0.178& 0.269	& 0.212& 0.327	& 0.198 & 0.300	& 0.189 &0.285	& 0.169 &0.268	& 0.167 &0.263\\
         
         &720 &  \textbf{0.197} & \textbf{0.285} &	0.212 & 0.310	& \underline{0.200} & \underline{0.292} & \underline{0.200} &0.295	& 0.225 & 0.317 & 0.245 &0.352 & 0.220 & 0.320 & 0.250& 0.341 &	0.204& 0.300 & 0.201& 0.295 \\

         \midrule
         
        \multirow{4}{*}{\rotatebox{90}{\emph{Solar}}}
         &96 & \underline{0.175} & \underline{0.228}	& \textbf{0.167} & \textbf{0.220} & 0.224 & 0.278	& 0.203 & 0.237 & 0.181 & 0.240 & 0.201 & 0.304 & 0.219 & 0.314 & 0.188 & 0.252	& 0.289 & 0.337	& 0.284 & 0.325\\
         
         &192 & \textbf{0.186} & \textbf{0.235} & \underline{0.187} & \underline{0.249} & 0.253 & 0.298	& 0.233 & 0.261	& 0.196 & 0.252	& 0.237 & 0.337	& 0.231 & 0.322	& 0.215 & 0.280	& 0.319 & 0.397	& 0.307 & 0.362\\
         
         &336 & \textbf{0.200} & \underline{0.246} & \textbf{0.200} & 0.258	& 0.273 & 0.306	& 0.248 & 0.273	& \underline{0.216} & \textbf{0.243}	& 0.254 & 0.362	& 0.246 & 0.337	& 0.222 & 0.267	& 0.352 & 0.415	& 0.333 & 0.384\\
         
         &720 & \textbf{0.203} & \textbf{0.249}	& \underline{0.215} & \underline{0.250} & 0.272 & 0.308	& 0.249 & 0.275	& 0.220 & 0.256	& 0.280 & 0.397	& 0.280 & 0.363	& 0.226 & 0.264	& 0.356 & 0.412	& 0.335 & 0.383\\

         \midrule
         
        \multirow{4}{*}{\rotatebox{90}{\emph{Traffic}}}
         &96 & 0.377 & \underline{0.256}	& \underline{0.366} & 0.259	& 0.383 &0.272	& 0.395& 0.268 & \textbf{0.360} & \textbf{0.249}	& 0.577& 0.360	& 0.593 & 0.321	& 0.413& 0.290 & 0.412& 0.286 & 0.649 &0.389\\
         
         &192 & 0.392 & 0.263 &	0.381 & 0.265 & \underline{0.380} & \underline{0.259} & 0.417& 0.276	& \textbf{0.379} & \textbf{0.256} & 0.607 &0.374 &	0.617 & 0.336	& 0.409& 0.289 &	0.424 &0.291	&0.601 &0.366  \\
         
         &336 &   0.406& \textbf{0.266} & \underline{0.397} & \underline{0.269} & 0.410 &0.285	&0.433& 0.283 &	\textbf{0.385} & 0.270	& 0.624& 0.384 &	0.629& 0.336	&0.425 &0.299 &	0.438& 0.299	& 0.609& 0.369\\
         
         &720 & 0.447& \underline{0.287} & \textbf{0.429} &0.292	&0.454 &0.313 &	0.467& 0.302	& \underline{0.430} & \textbf{0.281}	& 0.625 & 0.381 &	0.640 & 0.350	& 0.525 &0.373	& 0.467& 0.317	&0.647& 0.387\\

         \midrule
         
        \multirow{4}{*}{\rotatebox{90}{\emph{Weather}}}
         &96 & \textbf{0.143} & \textbf{0.188}	&0.147 & \underline{0.197}	&0.148 &0.200	&0.174 &0.214 &	\underline{0.145} & \underline{0.197}	& 0.221 &0.304 &	0.172& 0.220	&0.194& 0.234 &	0.175& 0.235 & 0.192 &0.232\\
         
         &192 & \textbf{0.186} & \textbf{0.230}	&0.189& 0.239	& 0.191& 0.241	&0.221 &0.254 &	\underline{0.187} & \underline{0.238}	&0.325 &0.372 &	0.219 & 0.261	& 0.229 &0.266	& 0.216 &0.274	& 0.240& 0.271 \\
         
         &336 & \textbf{0.236} & \textbf{0.269}	& 0.241 & \underline{0.280} & 0.240 & 0.281 & 0.278 &0.296 & \underline{0.240} & \underline{0.280} & 0.386 &0.408	& 0.280 & 0.306	& 0.266 &0.295	& 0.262& 0.314	& 0.292& 0.307 \\
         
         &720 & 0.314 & \textbf{0.326} &	\underline{0.310} & 0.330	& \textbf{0.307} & \underline{0.329} &	0.358& 0.347 &	0.315 & 0.330	& 0.415& 0.423 &	0.365 & 0.359	& 0.323 &0.340	& 0.327& 0.367 & 0.364& 0.353 \\
        \bottomrule
    \end{tabular} }
    \caption{Full result of multivariate long-term time series forecasting on eight datasets with various prediction lengths $T\in \{96, 192, 336, 720\}$ and fixed look-back window size $L=96$.}\label{tab:a1}
\end{table*}

\begin{table*}[t]
    \resizebox{\textwidth}{!}{
    \begin{tabular}{c|c|cc|cc|cc|cc|cc|cc|cc|cc}
        \toprule
        \multicolumn{2}{c|}{Models} & \multicolumn{2}{c|}{Dualformer} & \multicolumn{2}{c|}{Autoformer} & \multicolumn{2}{c|}{FEDformer} & \multicolumn{2}{c|}{FiLM} &
        \multicolumn{2}{c|}{PatchTST} &
        \multicolumn{2}{c|}{iTransformer} &
        \multicolumn{2}{c|}{FreTS} &
        \multicolumn{2}{c}{DLinear} \\
        \midrule
       \multicolumn{2}{c|}{Metric} & MSE & MAE & MSE & MAE &  MSE & MAE &  MSE & MAE &  MSE & MAE &  MSE & MAE &  MSE & MAE &  MSE & MAE \\
        \midrule
        \multirow{4}{*}{\rotatebox{90}{\emph{ETTh1}}}
         &96 & \textbf{0.054}  & \textbf{0.176}  & 0.088 & 0.238 & 0.084 & 0.220 & 0.057 & 0.182 & \textbf{0.054} & \underline{0.177} & \underline{0.055} & 0.181 & 0.071 & 0.206 & 0.056 & 0.180 \\
         &192 & \textbf{0.069}  & \textbf{0.202} & 0.097 & 0.239 &  0.108 & 0.249 & 0.072 & 0.207 & \underline{0.071} & 0.205 & \underline{0.071} & \underline{0.204} & 0.104 & 0.245 & 0.075 & 0.209  \\
         &336 & \textbf{0.077}  & \textbf{0.221} & 0.104 & 0.260 &  0.123 & 0.278 &  0.084 & 0.230 & 0.083 & 0.228 & \underline{0.081} & \underline{0.226} & 0.107 & 0.258 & 0.092 & 0.238   \\
         &720 & 0.090 & 0.237 & 0.132 & 0.290 &  0.146 & 0.303 & 0.091 & 0.239 & \underline{0.086} & \underline{0.234} & \textbf{0.080} & \textbf{0.226} & 0.126 & 0.283 & 0.168 & 0.334 \\
        \midrule
        \multirow{4}{*}{\rotatebox{90}{\emph{ETTh2}}}
         &96 & \textbf{0.126}  & 0.274  & 0.152 & 0.303 & 0.129 & \textbf{0.271} & \underline{0.128} & \underline{0.273} & 0.130 & 0.283 & 0.129 & 0.278 & \underline{0.128} & \textbf{0.271} & 0.132 & 0.280 \\
         &192 & \textbf{0.163}  & \textbf{0.319} & 0.191 & 0.340 &  0.187 & 0.330 & 0.189 & 0.343 & \underline{0.169} & 0.328 & \underline{0.169} & \underline{0.324} & 0.185 & 0.330 & 0.176 & 0.329  \\
         &336 & \textbf{0.188}  & \textbf{0.351} & 0.233 & 0.380 &  0.231 & 0.378 &  0.201 & 0.364 & \underline{0.193} & 0.358 & 0.194 & \underline{0.355} & 0.231 & 0.378 & 0.210 & 0.369   \\
         &720 & \textbf{0.216} & \textbf{0.373} & 0.260 & 0.404 &  0.278 & 0.422 & 0.224 & 0.380 & \underline{0.221} & \underline{0.379} & 0.225 & 0.381 & 0.268 & 0.409 & 0.290 & 0.438 \\
         \midrule
        \multirow{4}{*}{\rotatebox{90}{\emph{ETTm1}}}
         &96 & \textbf{0.026}  & \textbf{0.121}  & 0.050 & 0.174 & 0.034 & 0.143 & 0.029 & 0.128 & \textbf{0.026}  & \textbf{0.121} & \textbf{0.026} & \underline{0.122} & 0.033 & 0.140 & \underline{0.028} & 0.125 \\
         &192 & \textbf{0.039}  & \underline{0.150} & 0.110 & 0.250 &  0.066 & 0.203 & \underline{0.041} & 0.154 & \textbf{0.039}  & \underline{0.150} & \textbf{0.039} & \underline{0.149} & 0.058 & 0.186 & 0.043 & 0.154  \\
         &336 & \textbf{0.052}  & \underline{0.173} & 0.085 & 0.236 &  0.071 & 0.210 &  \underline{0.053} & 0.174 & \underline{0.053} & \underline{0.173} & \textbf{0.052} & \textbf{0.172} & 0.071 & 0.209 & 0.062 & 0.183   \\
         &720 & \textbf{0.070} & \textbf{0.200} & 0.120 & 0.283 &  0.109 & 0.259 & \underline{0.071} & \underline{0.205} & 0.073 & 0.206 & 0.073 & 0.207 & 0.102 & 0.248 & 0.080 & 0.211 \\
         \midrule
        \multirow{4}{*}{\rotatebox{90}{\emph{ETTm2}}}
         &96 & \textbf{0.062}  & \textbf{0.182}  & 0.098 & 0.241 & 0.066 & 0.197 & 0.066 & 0.191 & 0.065 & 0.186 & \underline{0.063} & \textbf{0.182} & \underline{0.063} & 0.189 & 0.064 & \underline{0.184} \\
         &192 & \underline{0.091}  & \underline{0.224} & 0.164 & 0.315 &  0.104 & 0.250 & 0.096 & 0.235 & 0.094 & 0.231 & \textbf{0.090} & \textbf{0.223} & 0.102 & 0.245 & 0.092 & 0.227  \\
         &336 & \textbf{0.117}  & \underline{0.260} & 0.178 & 0.325 &  0.155 & 0.303 &  0.123 & 0.269 & \underline{0.120} & 0.265 & \textbf{0.117} & \textbf{0.259} & 0.130 & 0.279 & 0.122 & 0.265   \\
         &720 & \textbf{0.170} & \underline{0.321} & 0.189 & 0.340 &  0.193 & 0.343 & 0.173 & 0.323 & \underline{0.172} & 0.322 & 0.175 & \textbf{0.320} & 0.178 & 0.325 &  0.174 & \textbf{0.320} \\
        \bottomrule
    \end{tabular}
    }
    \caption{Full result of univariate long-term time series forecasting on ETT datasets with various prediction lengths $T\in \{96, 192, 336, 720\}$ and fixed look-back window size $L=96$.}\label{tab:a2}
\end{table*}
\subsection*{Model robustness}
All experiments were repeated three times using three different random seeds to ensure statistical robustness. We report the mean and standard deviation (Mean ± STD) for Dualformer and three other models (TimeMixer, PatchTST, and iTransformer). As shown in Table~\ref{tab:error_bar}, our Dualformer model consistently achieves strong and stable performance across various datasets and prediction lengths, highlighting both its effectiveness and reliability.

\subsection*{Normalized vs. Re-normalized evaluation}

Previous work in long-term time series forecasting (LTSF) typically reports MAE and MSE on z-score normalized data~\citep{zhou2021informer, wu2021autoformer, zhou2022fedformer}. However, these absolute metrics are affected by the original data scale, making the results difficult to interpret. Normalized evaluation can produce deceptively low error values, which may not reflect real-world performance. To address this issue, we further assessed model predictions on re-normalized data using MAE, RMSE, and WAPE.


Table~\ref{tab:02} presents results on re-normalized data across four datasets. Compared to Table~\ref{tab:01}, it shows that errors appear larger when evaluated on the original scale. Despite this, our model demonstrates stable performance across all datasets. Notably, the consistently low WAPE values indicate that Dualformer generalizes well and remains robust to varying data distributions.

\subsection*{Look-back window analysis}
To evaluate the impact of the input window size on forecasting performance, we conducted experiments by varying the look-back window size 
$L\in \{96,192,336,720\}$, while keeping the prediction length $T=336$. As shown in Fig.~\ref{fig:window}, longer input consistently improved the forecasting accuracy on both ETTh1 and Weather datasets. This highlights the benefit of using extended temporal context in long-term forecasting tasks.

\subsection*{Full Results}
Here we show the full results for multivariate forecasting (Table \ref{tab:a1}) and univariate forecasting (Table \ref{tab:a2}). The \textbf{best} and \underline{second-best} results are highlighted. Our dualformer model consistently outperforms selected baselines on most datasets across different prediction lengths.

\end{document}